\documentclass{article}
\usepackage{ijcai25}
\usepackage[utf8]{inputenc}  

\usepackage[hidelinks]{hyperref} 
\usepackage{url}  
\usepackage{times}  
\usepackage[small]{caption} 
\usepackage{lipsum}  
\usepackage{bbm}  

\usepackage[switch]{lineno}  

\usepackage{booktabs}  
\usepackage{multirow}  
\usepackage{pifont}  
\usepackage{colortbl}  
\usepackage{xcolor}  
\usepackage{threeparttable}  
\usepackage{makecell}  
\usepackage{tabularx}  

\usepackage{graphicx}  
\usepackage{subfigure}  
\usepackage{wrapfig}  
\usepackage{subcaption}  
\usepackage{float}  
\usepackage{stfloats}  

\usepackage{pifont}  
\usepackage{textcomp}  

\usepackage{dsfont}  

\usepackage{amsmath}  
\usepackage{amsthm}  
\usepackage{amsfonts}  
\usepackage{amssymb}  
\usepackage{mathrsfs}  
\usepackage{bm}  
\usepackage{nicefrac}  

\usepackage{bbm}  
\usepackage{algorithm}  
\usepackage{algorithmic}  

\def\eqref#1{equation~\ref{#1}}

\def\1{\bm{1}}

\DeclareMathAlphabet{\mathsfit}{\encodingdefault}{\sfdefault}{m}{sl}
\SetMathAlphabet{\mathsfit}{bold}{\encodingdefault}{\sfdefault}{bx}{n}

\title{MedualTime: A Dual-Adapter Language Model for Medical Time Series-Text Multimodal Learning}

\author{
Jiexia Ye$^1\dagger$\and
Weiqi Zhang$^2\dagger$\and
Ziyue Li$^{3*}$\and
Jia Li$^{2*}$ \and
Meng Zhao$^{4}$ \and
Fugee Tsung$^2$ 
\affiliations
$^1$The Hong Kong University of Science and Technology (Guangzhou) \\
$^2$The Hong Kong University of Science and Technology\\
$^3$Technical University of Munich \\
$^4$Columbia University\\
\emails
jye324@connect.hkust-gz.edu.cn,
wzhangcd@connect.ust.hk,
ziyue.li@tum.de,
mz3048@cumc.columbia.edu,
\{jialee, season\}@ust.hk
}

\begin{document}

\maketitle


\begin{abstract}

The recent rapid advancements in language models (LMs) have garnered attention in medical time series-text multimodal learning.
However, existing contrastive learning-based and prompt-based LM approaches tend to be biased, often assigning a primary role to time series modality while treating text modality as secondary. We classify these approaches under a temporal-primary paradigm, which may overlook the unique and critical task-relevant information embedded in text modality like clinical reports, thus failing to fully leverage mutual benefits and complementarity of different modalities.
To fill this gap, we propose a novel textual-temporal multimodal learning paradigm that enables either modality to serve as the primary while being enhanced by the other, thereby effectively capturing modality-specific information and fostering cross-modal interaction. In specific, we design \textit{\textbf{MedualTime}}, a language model composed of dual adapters to implement temporal-primary and textual-primary modeling simultaneously. Within each adapter, lightweight adaptation tokens are injected into the top layers of LM to encourage high-level modality fusion. The shared LM pipeline by dual adapters not only achieves adapter alignment but also enables efficient fine-tuning, reducing computational resources. Empirically, MedualTime demonstrates superior performance on medical data, achieving notable improvements of 8\% accuracy and 12\% F1 in supervised settings. 
Furthermore, MedualTime's transferability is validated by 
few-shot transfer experiments from coarse-grained to fine-grained medical data.

\end{abstract}

\section{Introduction}
\label{sec:int}

In the medical domain, time series and text modalities are pivotal in representing patient information. Time series, such as electrocardiograms (ECG), electroencephalograms (EEG), and vital signs, provide continuous measurements that reflect a patient’s physiological states over time \cite{wang2024contrast}. On the other hand, text modalities encompass various clinical documentation, such as radiology reports and physicians' notes, which offer contextual insights for understanding the broader scope of a patient’s clinical condition \cite{davuluri2024overview}. Traditional methods typically involve designing models tailored for single data types and tasks \cite{fries2021ontology,9845479,weimann2021transfer}.
However, jointly modeling time series with text modalities offers richer insights for clinical decision-making.
For instance, EEG signals capture electrical activity in the brain, while clinical reports usually provide health history. Analyzing only the symptom history might suggest epilepsy but struggles to specify seizure types, while EEGs detect abnormal brain activity but lack individual medical context. Integrating both modalities can improve diagnostic precision and rationality. A key challenge for jointly modeling is to effectively represent and exploit the complementarity and interactions between different modalities \cite{guo2019deep}.

Recently, large-scale pre-trained language models (LMs) have shown exceptional proficiency in understanding sequential data \cite{LLM4TS,LLMTIME}, sparking interest in leveraging them to integrate time series and text modalities. Several contrastive learning-based works in medical domain utilize language models as textual encoders to extract meaningful representations from clinical knowledge, guiding the pre-training of time series encoder \cite{liuzero,yu2024ecg,king2023multimodal}. However, the clinical knowledge is not present during the inference stage, leading to a potential loss of critical information. For instance, METS \cite{li2024frozen} utilizes a frozen LM to derive embeddings from clinical reports,  aligning them with ECG embedding through contrastive learning to augment ECG representation. And only ECG encoder provides decision for inference. Other prompt-based works integrate the text modality into a prompt to guide LM to comprehend time series input \cite{jia2024gpt4mts,liu2023unitime}. Specifically, MedTsLLM \cite{chan2024medtsllm} augments the text prompt with patient-specific clinical information to adapt LLM for medical signal processing.


\begin{figure}
  \centering
  
  \includegraphics[width=0.37\textwidth]
  {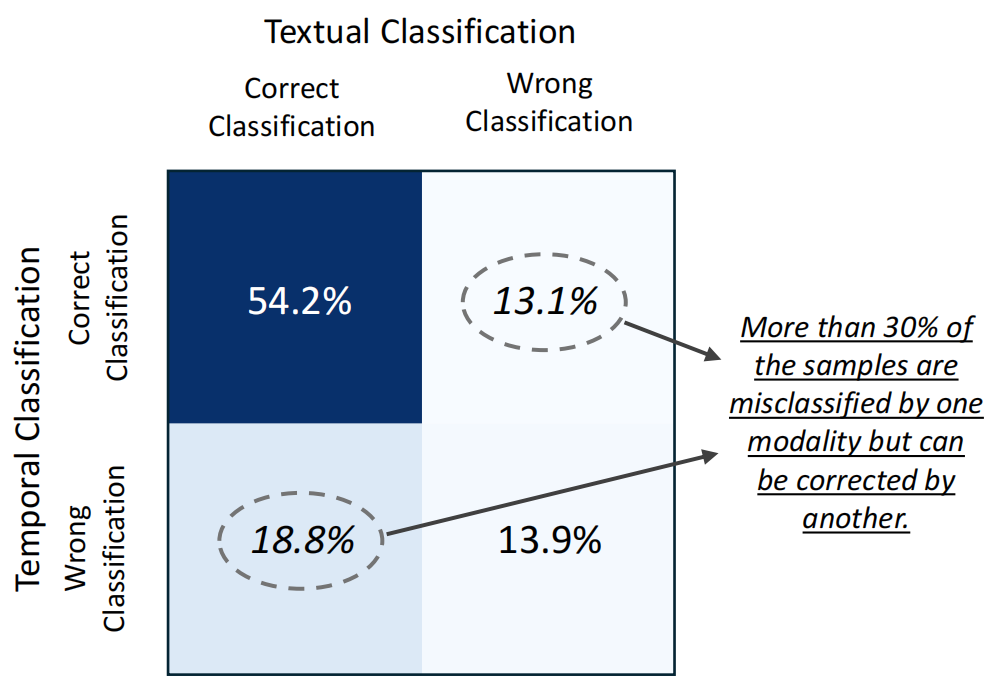}
  \caption{Unimodal classification results on the ECG multimodal dataset (PTB-XL 5 Classes), using LSTM for time series classification and BERT for text classification. The circled samples are misclassified by one modality but corrected by another, demonstrating the complementary information of different modalities.}
  \label{fig:par}
  \vspace{-3mm}
\end{figure}

\begin{figure*}[!t] 
  \centering
  
  \includegraphics[width=0.99\textwidth]{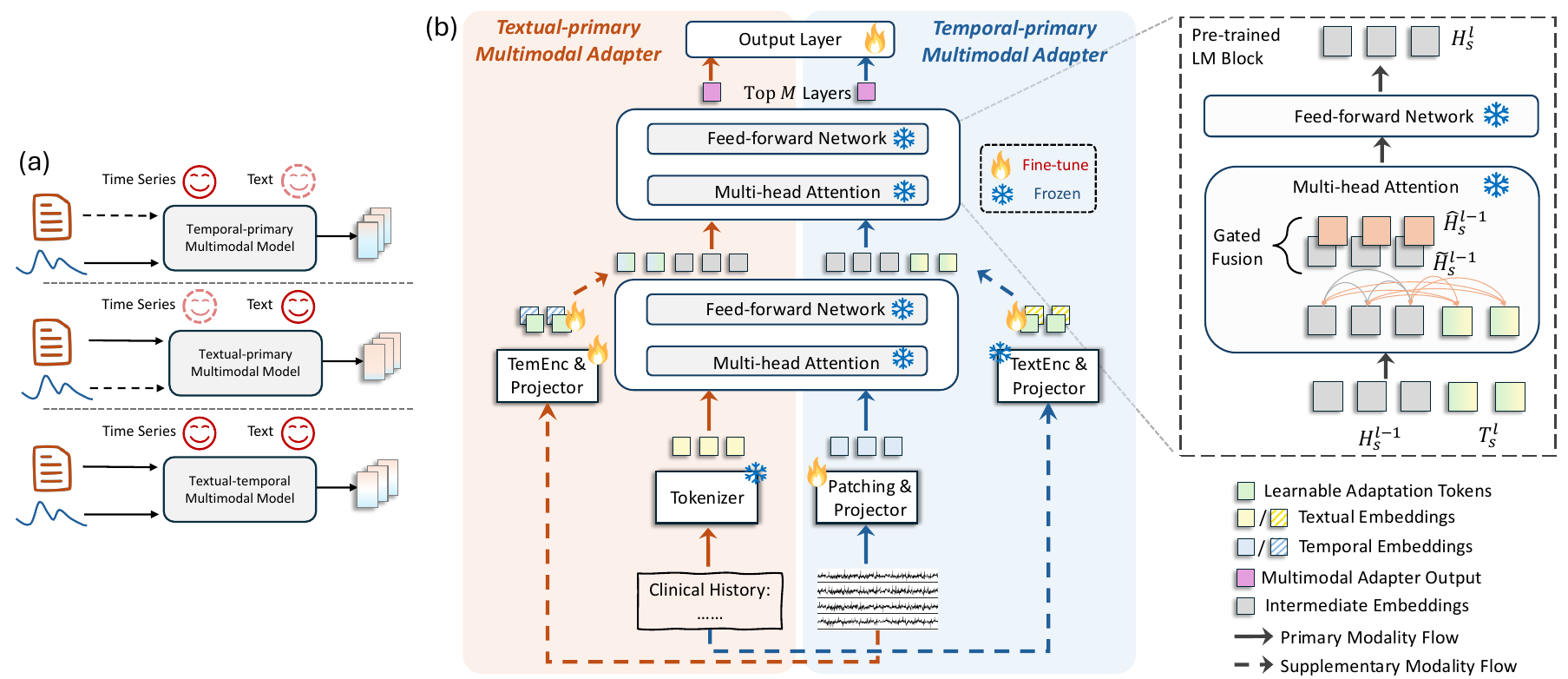}
  
  \caption{
  (a) Comparisons between different time series multimodal modeling paradigms. The solid line represents the primary modality and the dashed line for the secondary modality. (b) MedualTime architecture. It consists of dual adapters to model time series and text as primary modality respectively. Dual adapters share the same LM parameters to reduce computational cost and realize adapter alignment. The LM's pre-trained knowledge is preserved by adopting a zero-initialized gating strategy. The cross-modal fusion is achieved by injecting trainable adaptation tokens in the top layers of LM within each adapter.}

  \label{fig:mod}
\end{figure*}

In these LM-based multimodal time series-text works, time series is typically considered as the primary modality, being more relevant for decision-making, while text serves as an auxiliary modality to enhance the time series embedding, either by projecting textual knowledge into the time series encoder using contrastive learning \cite{liuzero} or by guiding LM with a textual prompt to comprehend temporal inputs \cite{chan2024medtsllm}. We classify these approaches as temporal-primary multimodal models (Figure \ref{fig:mod} (a)). However, in some cases, the textual information is no less important than temporal information. As shown in Figure \ref{fig:par}, we conduct a unimodal classification experiment on the PTB-XL ECG dataset and find that 18.8\% of samples are correctly classified by the text modality but misclassified by the time series modality, while 13.1\% shows the reverse. This highlights the complementarity of these two modalities and suggests that the text modality contains even more unique task relevant information in some cases. Therefore, viewing text modality as auxiliary may introduce bias and fail to capture essential information while a text-primary perspective could enable a more comprehensive understanding of the informative textual content.

To fully exploit the complementarity and mutual benefits of different modalities, we propose a novel textual-temporal multimodal learning paradigm to integrate both temporal-primary and textual-primary perspectives (Figure \ref{fig:mod} (a)). However, to effectively construct LM-based approach of such paradigm is technically non-trivial. The most straightforward solution is to train two LM-based submodels separately. Nevertheless, there remain two-fold challenges: First, considering LMs involved, two separately trained submodels suffer non-negligible computational costs. Second, the integration and design of submodels should fuse the two modalities from different perspectives to capture both shared and unique information effectively. Note that the naive multimodal concatenation at LM input layer of existing works \cite{chan2024medtsllm} is difficult to extract the intertwined and high-level multimodal semantics.

To address the aforementioned challenges, we propose MedualTime, a language model for medical time series-text multimodal learning,
consisting of a temporal-primary multimodal adapter and a textual-primary multimodal adapter to effectively explore the complementary information in multimodal medical input.
Under dual adapter design, each modality has the chance to serve as the primary modality and get improved by the other modality. Within each adapter, multimodal fusion is achieved by injecting learnable adaptation tokens into the top layers to extract high-level multimodal semantics.
Furthermore, both adapters share the same LM backbone to reduce computational resources. Meanwhile, we keep the majority of LM's parameters frozen to make different modalities benefit from its pre-trained knowledge. We update only a small portion of LM's parameters, adapting it to our task while enabling efficient fine-tuning. In addition, by pipeline sharing, the modality alignment of different adapters could be accomplished.  Our main contributions are as below:

\begin{itemize}

\item
We are the first to propose a textual-temporal multimodal learning paradigm  that treats both time series and text modality equally in medical scenarios. This paradigm fully leverages the rich complementary semantics and captures the intricate interaction between different modalities.


\item
We propose \textbf{MedualTime}, a dual-adapter language model for medical time series-text multimodal learning. Each adapter performs the mutual integration of time series and text modalities by introducing learnable tokens into the top layers of the LM backbone, facilitating high-level multimodal semantic fusion. The frozen shared LM pipeline allows both adapters to leverage the pre-trained knowledge and achieve alignment in the same semantic space, largely reducing the computational resources with adapter tuning.


      
\item
\textbf{MedualTime} demonstrates superior performance on public medical multimodal datasets, showing its strong generalization and transferability in medical diagnosis tasks. Notably, it achieves an average improvement of \textbf{8\%} in accuracy and \textbf{12\%} in F1 score under supervised learning. The code is in \url{https://github.com/start2020/MedualTime}.


\end{itemize}






\section{Related Work}
\label{sec:rel}
\label{sec:Related_work}
In this section, we discuss LM based approaches for time series-text multimodal learning in both medical and general domains. We categorize them into two branches based on how they derive multimodal representation.


\textbf{Contrastive Learning based Approaches} adopt contrastive learning to align time series and text modalities into separate but coordinated spaces to enforce shared information between modalities \cite{liang2024foundations}. This branch includes METS \cite{li2024frozen}, MERL \cite{liuzero}, ESI \cite{yu2024ecg} and \cite{king2023multimodal}. They leverage LMs to obtain embedding representations of the text modality, which then guide the pre-training of time series encoder, enhancing the quality and robustness of time series representation.
For instance, MERL employs contrastive learning to enhance ECG signal representations under the supervision of clinical reports. However, during training, contrastive learning tends to focus on capturing shared semantics between modalities, often overlooking modality-specific information. Furthermore, in the inference stage, only the time series modality is available, while the text modality is absent. As a result, this framework relies solely on time series data for decision-making, potentially disregarding unique and task-relevant information embedded in the text. This limitation can lead to suboptimal model performance.

\textbf{Prompt-based Approaches} incorporate text modality into a textual prompt, which is processed by a language model to generate embeddings. Such embedding is then prepended as a prefix of temporal embedding to enhance LLM's reasoning capacities on time series modality. This branch includes Time-LLM \cite{jin2023timellm}, UniTime \cite{liu2023unitime}, GPT4MTS \cite{jia2024gpt4mts}, InstructTime \cite{cheng2024advancing}, MedTsLLM \cite{chan2024medtsllm}. For instance, Time-LLM assembles dataset description, task instruction, and data statistics into a textual prompt. The prompt is turned to textual embedding by a frozen LLM, concatenated before the temporal embedding as a prefix to facilitate LLM’s understanding of time series data. However, such sequential concatenation implies that the concatenated modalities are not equally important, making LLM focus more on time series.

All these LM-based multimodal works consider time series as the primary modality for decision-making, with text serving as an auxiliary modality to enhance time series modeling. In contrast, MedualTime allows each modality to act as a primary modality, which can comprehensively capture the unique and shared semantics in different modalities.

\section{Methodology}
\label{sec:met}

  

In this work, we focus on sample-level medical time series-text data.
Specifically, each sample is a time-text pair (e.g., ECG signal and its coupled clinical report). The whole dataset is denoted as $\mathcal{S}=\left\{\left(\bm{X_1}, \bm{S_1}\right),\left(\bm{X_2}, \bm{S_2}\right),...,\left(\bm{X_N}, \bm{S_N}\right)\right\}$, where $\bm{X_i}\in \mathbb{R}^{T\times d}$ denotes a $d$-dimension multivariate time series modality with length $T$ and $\bm{S_i}$ denotes the paired textual modality. For simplicity, we omit the sample indicator subscript in the following. To fully utilize the complementary information of different modalities, MedualTime consists of two multimodal adapters, namely a textual-primary multimodal adapter and a temporal-primary multimodal adapter. Each adapter treats one modality as the primary modality and enhances it with the other modality. Both adapters share the same frozen pre-trained language model with $L$ layers. Each adapter implements multimodal fusion in the topmost $M~(M\leq L)$ transformer blocks of the language model. The shared language model backbone facilitates efficient fine-tuning and encourages the dual adapters' embedding space alignment.

\subsection{Textual-primary Multimodal Adapter}
\label{sec:adapter1}


  

Processed by the textual tokenizer, the text input can be modeled by $I^s$-length word tokens with embedding $\bm{E}_s \in \mathbb{R}^{I^s\times D}$, where $D$ is the hidden dimension. 
For the first $L-M$ transformer layers, 
which are standard transformer layers, the forward process of layer-$l$ is:
\begin{equation}
\resizebox{\columnwidth}{!}{%
$\tilde{\bm{H}}_{s}^{l-1}  =\operatorname{LN}\left(\operatorname{MHA}\left(\bm{W}_{q}^{l}\bm{H}_{s}^{l-1}, \bm{W}_{k}^{l}\bm{H}_{s}^{l-1}, \bm{W}_{v}^{l}\bm{H}_{s}^{l-1}\right)\right)+\bm{H}_{s}^{l-1}$%
}
\label{eq:forward1}
\end{equation}
\begin{equation}
\bm{H}_{s}^l  =\operatorname{LN}\left(\operatorname{MLP}\left(\tilde{\bm{H}}_{s}^{l-1}\right)\right)+\tilde{\bm{H}}_{s}^{l-1}
\label{eq:forward2}
\end{equation}
where $\bm{H}_{s}^l$ is the output of layer-$l$ with $\bm{H}_{s}^0 = \bm{E}_s$, $\operatorname{MHA}, \operatorname{LN}, \operatorname{MLP}$ denote the multi-head attention, the layer normalization, and the multi-layer perception, respectively. To obtain the query, key, value matrics at layer-$l$, $\bm{W}_{q}^{l}, \bm{W}_{k}^{l}, \bm{W}_{v}^{l}$ are parameterized by the pre-trained language model. Meanwhile, the attention operation  $\operatorname{Attention}$ is defined by:
\begin{equation}
\operatorname{Attention}\left(\bm{Q}, \bm{K}, \bm{V}\right)=\operatorname{softmax}\left(\bm{Q}\bm{K}^T/\sqrt{d_k}\right) \bm{V}
\end{equation}
where $\bm{Q}, \bm{K}, \bm{V}$ are corresponding query, key, and value matrices, $d_k$ is the dimension of key.

Furthermore, inspired by the adapter architecture in \cite{zhang2023llama}, we utilize a lightweight adapter mechanism to achieve multimodal modeling at the topmost $M$ transformer blocks. Specifically, we adopt learnable length-$P$ adaptation tokens $\bm{T}_s^l$ at each multimodal fusion layer $l~(L-M+1\leq l \leq L)$ 
, where the adaptation tokens $\bm{T}_s^l\in \mathbb{R}^{P\times D}$ have the same dimension as language model. As to the secondary temporal modality, a trainable temporal encoder and a cross-modal projector are utilized to transform the time series input into the language model embedding space:
\begin{equation}
    \bm{Z}_s = \operatorname{Projector}\left(\operatorname{TemEncoder}\left(\bm{X}\right)\right)
    \label{eq:temenc}
\end{equation}
The temporal encoder can be any time-series encoder that best fits the specific datasets, while the projector is a linear layer responsible for dimension transformation.
For decreasing the computational cost, different multimodal fusion layers share the same temporal embedding. Thus, the adaptation tokens of textual-primary multimodal adapter is calculated by: 
\begin{equation}
     \tilde{\bm{T}}_s^l = \bm{T}_s^l + \bm{Z}_s
     \label{eq:adaptation}
\end{equation}

For the topmost $M$ transformer layers, the multimodal forward process is formalized as:
\begin{equation}
\resizebox{\columnwidth}{!}{%
$
\tilde{\bm{H}}_{s}^{l-1}  = \operatorname{LN}\left(\operatorname{MHA}\left(\bm{W}_{q}^{l}\bm{H}_{s}^{l-1}, \bm{W}_{k}^{l}\bm{H}_{s}^{l-1}, \bm{W}_{v}^{l}\bm{H}_{s}^{l-1}\right)\right)+\bm{H}_{s}^{l-1}
$
}
\end{equation}
\begin{equation}
\resizebox{\columnwidth}{!}{%
$
\hat{\bm{H}}_{s}^{l-1}  = \operatorname{LN}\left(\operatorname{MHA}\left(\bm{W}_{q}^{l}\bm{H}_{s}^{l-1}, \bm{W}_{k}^{l}\tilde{\bm{T}}_s^l, \bm{W}_{v}^{l}\tilde{\bm{T}}_s^l\right)\right)+\bm{H}_{s}^{l-1}
$
}
\label{eq:mha}
\end{equation}
\begin{equation}
\resizebox{\columnwidth}{!}{%
$
\bm{H}_{s}^l  = \operatorname{LN}\left(\operatorname{MLP}\left(\textit{Gate}^l\hat{\bm{H}}_{s}^{l-1}+\tilde{\bm{H}}_{s}^{l-1}\right)\right)+\left(\textit{Gate}^l\hat{\bm{H}}_{s}^{l-1}+\tilde{\bm{H}}_{s}^{l-1}\right)
$
}
\label{eq:gate}
\end{equation}


In particular, combined with the pre-trained projection matrices $\bm{W}_k^l, \bm{W}_v^l$, the learnable adaptation tokens serve as key, value matrices of the multi-head attention layer. In Equation (\ref{eq:gate}), we perform a zero-initialized gating strategy to achieve multimodal adaptation token fusion \cite{zhang2023llama}. Gating parameter $\textit{Gate}^l$ will be initialized as zero at the beginning of training, the multimodal adaptation tokens will be injected gradually, which can preserve the pre-trained knowledge and capacities of LMs.

\subsection{Temporal-primary Multimodal Adapter}

Considering the sequential property of time series, the temporal-primary multimodal adapter takes the time series data as the language model input. 
We utilize the common patching strategy for time series modeling in related works \cite{nie2022patchtst,zhou2024one}.
Several adjacent timestamps will be assembled as a token, which can provide local semantic information within a time series. For a pre-defined patch size $p$ and stride $s$, the time series input $\bm{X} \in \mathbb{R}^{T\times d}$ can be reorganized as $\tilde{\bm{X}} \in \mathbb{R}^{T_{s}\times (p\times d)}$, where $T_s=\left\lceil\frac{T-p}{s}\right\rceil+1$ is the number of temporal tokens. Subsequently, we utilize a projector (i.e. linear layer) to adjust the dimension of temporal tokens. The adjusted temporal token can be denoted as $\bm{E}_t ~(\bm{E}_t \in \mathbb{R}^{T_s\times D})$.
With $\bm{H_t^0}=\bm{E}_t$ as the input of the first transformer layer, the model forward process will be similar to the ones introduced in Section \ref{sec:adapter1}, e.g., Equation (\ref{eq:forward1}-\ref{eq:forward2}) and Equation (\ref{eq:adaptation} - \ref{eq:gate}).

For the secondary text input, we use a frozen pre-trained language model (LM) as a text encoder (similar to the temporal encoder in Equation (\ref{eq:temenc})) to extract textual information:
\begin{equation}
    \bm{Z}_t = \operatorname{Projector}\left(\operatorname{LM}\left(\bm{S}\right)\right)
\end{equation}

\subsection{Pre-trained Language Model Parameters Sharing}
Aided by our dual-adapter model design, most of the parameters in the pre-trained language model (e.g., the attention weight matrices $\bm{W}_q, \bm{W}_k, \bm{W}_v$, and the MLP layer of each transformer block) can be shared between the textual-primary and temporal-primary multimodal adapters. On the one hand, freezing these parameters preserves the language model's knowledge and sequential modeling capabilities. On the other hand, since most of the parameters in our proposed adapters are shared, the increase in training parameters is minimal compared to using a single adapter. This ensures complementary modeling between the two modalities while maintaining efficient fine-tuning. Additionally, by sharing the same LM pipeline, the embedding spaces of the different adapters align easily, further facilitating the integration of dual adapters.

\subsection{Training Loss}
\label{sec:training}
\textbf{Supervised Learning.} For supervised classification, we add the last transformer layer output of each adapter together to obtain the final multimodal representation. Then, an extra linear classifier and the cross-entropy loss are used for supervised training.

\textbf{Unsupervised Representation Learning.} 
For unsupervised representation learning, following previous works \cite{zhang2023co}, we adopt the contrastive learning paradigm. In particular, for data augmentation, we add random Gaussian noise to the original input. The noise-corrupted sample and its original sample are a positive pair within each adapter.
We denote $\bm{H}_s^{'L}$ as the augmentation of $ \bm{H}_s^L$, and $\bm{H}_t^{'L}$ as the augmentation of $ \bm{H}_t^L$. 
The contrastive loss could be divided into two parts: within-adapter contrastive loss and cross-adapter contrastive loss.


Formally, by maximizing the agreement between positive pairs and minimizing the similarity between negative pairs (i.e., different input instances), in a mini-batch with size $B$, the within-adapter contrastive losses are
\begin{equation}
\resizebox{\columnwidth}{!}{$
\mathcal{L}_{s} = -\sum_{i=1}^{B}\log \frac{\exp \left(\operatorname{sim}\left(\bm{H}_{s,i}^{L}, \bm{H}_{s,i}^{'L}\right) / \tau\right)}{\sum_{k=1}^{B} \mathds{1}_{[k \neq i]} \exp \left(\operatorname{sim}\left(\bm{H}_{s,i}^{L}, \bm{H}_{s,k}^{L}\right) / \tau\right)}
$}
\label{eq:loss_source}
\end{equation}
\begin{equation}
\resizebox{\columnwidth}{!}{$
\mathcal{L}_{t} = -\sum_{i=1}^{B}\log \frac{\exp \left(\operatorname{sim}\left(\bm{H}_{t,i}^{L}, \bm{H}_{t,i}^{'L}\right) / \tau\right)}{\sum_{k=1}^{B} \mathds{1}_{[k \neq i]} \exp \left(\operatorname{sim}\left(\bm{H}_{t,i}^{L}, \bm{H}_{t,k}^{L}\right) / \tau\right)}
$}
\label{eq:loss_target}
\end{equation}


where $\mathds{1}_{[k \neq i]}$ is the indicator function and $\tau$ is the temperature parameter, $\operatorname{sim}(\cdot, \cdot)$ is the dot product between two $\ell_2$-normalized vectors.

The cross-adapter contrastive learning assumes that the embeddings from two adapters for one temporal-textual input pair should be similar. Concurrently, embeddings from different instances are considered negative pairs. In this vein, the cross-adapter contrastive loss is given by:
\begin{equation}
\resizebox{\columnwidth}{!}{%
$\begin{aligned}
\mathcal{L}_{cross} = -\sum_{i=1}^{B} \bigg( 
& \log \frac{\exp \left(\operatorname{sim}\left(\bm{H}_{s,i}^{L}, \bm{H}_{t,i}^{L}\right) / \tau\right)}{\sum_{k=1}^{B} \mathds{1}_{[k \neq i]} \exp \left(\operatorname{sim}\left(\bm{H}_{s,i}^{L}, \bm{H}_{t,k}^{L}\right) / \tau\right)} \\
& + 
\log \frac{\exp \left(\operatorname{sim}\left(\bm{H}_{t,i}^{L}, \bm{H}_{s,i}^{L}\right) / \tau\right)}{\sum_{k=1}^{B} \mathds{1}_{[k \neq i]} \exp \left(\operatorname{sim}\left(\bm{H}_{t,i}^{L}, \bm{H}_{s,k}^{L}\right) / \tau\right)}
\bigg)
\end{aligned}$%
}
\label{eq:loss_cross}
\end{equation}  


The overall loss function of unsupervised representation learning is given by:
\begin{equation}
    \mathcal{L}_{unsup} = \mathcal{L}_{s} + \mathcal{L}_{t} +\mathcal{L}_{cross}
\end{equation}


Note that for the variants of MedualTime, namely MedualTime (Time) and MedualTime (Text), we only adopt the within-adapter contrastive loss for training.

\section{Experiments}
\label{sec:exp}





\begin{table*}[tb]

\resizebox{\textwidth}{!}{

\begin{threeparttable}

\begin{tabular}{c|c|cccc|cccc|cccc|cccc|cc}
\toprule

\multicolumn{1}{c|}{}                                                               & \multicolumn{1}{c|}{}                                 & \multicolumn{8}{c|}{\textbf{PTB-XL}}                                                                                                                                                                                                                                                                                          & \multicolumn{8}{c|}{\textbf{TUSZ}}                                                                                                                                                                                                                                                                                            & \multicolumn{2}{c}{}                                                          \\ \cmidrule{3-18}
\multicolumn{1}{c|}{}                                                               & \multicolumn{1}{c|}{}                                 & \multicolumn{4}{c|}{\textbf{4 Classes}}                                                                                                                        & \multicolumn{4}{c|}{\textbf{5 Classes}}                                                                                                                  & \multicolumn{4}{c|}{\textbf{2 Classes}}                                                                                                                        & \multicolumn{4}{c|}{\textbf{4 Classes}}                                                                                                                  & \multicolumn{2}{c}{\multirow{-2}{*}{\textbf{Average}}}                        \\
\multicolumn{1}{c|}{\multirow{-3}{*}{\textbf{Modality}}}                            & \multicolumn{1}{c|}{\multirow{-3}{*}{\textbf{Model}}} & \textbf{Acc.}                         & \textbf{Pre.}                         & \textbf{Rec.}                         & \textbf{F1}                           & \textbf{Acc.}                         & \textbf{Pre.}                         & \textbf{Rec.}                         & \multicolumn{1}{c|}{\textbf{F1}}      & \textbf{Acc.}                         & \textbf{Pre.}                         & \textbf{Rec.}                         & \textbf{F1}                           & \textbf{Acc.}                         & \textbf{Pre.}                         & \textbf{Rec.}                         & \multicolumn{1}{c|}{\textbf{F1}}      & \textbf{Acc.}                         & \textbf{F1}                           \\ \midrule

                                                                                   & \textbf{LSTM}                                    & 0.68                                  & 0.60                                  & 0.48                                  & 0.48                                  & 0.67                                  & 0.63                                  & 0.50                                  & 0.52                                  & 0.76                                  & 0.53                                  & 0.54                                  & 0.54                                  & 0.58                                  & 0.44                                  & 0.27                                  & 0.26                                  & 0.67                                  & 0.45                                  \\
                                                                                   & \textbf{TimesNet}                                & 0.68                                  & 0.46                                  & 0.46                                  & 0.45                                  & 0.67                                  & 0.59                                  & 0.48                                  & 0.50                                  & 0.74                                  & 0.59                                  & 0.63                                  & 0.59                                  & 0.76                                  & 0.75                                  & \underline{ 0.72}                            & \underline{ 0.71}                            & 0.71                                  & 0.56                                  \\
                                                                                   & \textbf{LightTS}                                 & 0.68                                  & 0.59                                  & 0.53                                  & 0.54                                  & 0.59                                  & 0.46                                  & 0.44                                  & 0.45                                  & 0.74                                  & 0.53                                  & 0.53                                  & 0.54                                  & 0.71                                  & 0.72                                  & 0.58                                  & 0.58                                  & 0.68                                  & 0.53                                  \\
                                                                                   & \textbf{Dlinear}                                 & 0.68                                  & 0.58                                  & 0.50                                  & 0.49                                  & 0.61                                  & 0.46                                  & 0.41                                  & 0.41                                  & 0.78                                  & 0.52                                  & 0.52                                  & 0.52                                  & 0.71                                  & 0.62                                  & 0.60                                  & 0.59                                  & 0.70                                  & 0.50                                  \\
                                                                                   & \textbf{Pyraformer}                              & 0.76                                  & 0.66                                  & 0.59                                  & 0.58                                  & 0.66                                  & 0.56                                  & 0.49                                  & 0.51                                  & \underline{ 0.84}                            & 0.47                                  & 0.50                                  & 0.47                                  & \underline{ 0.75}                            & \underline{ 0.77}                            & 0.67                                  & \underline{ 0.72}                            & \underline{ 0.75}                            & 0.57                                  \\
                                                                                   & \textbf{ETSformer}                               & 0.72                                  & 0.63                                  & 0.57                                  & 0.55                                  & 0.54                                  & 0.45                                  & 0.38                                  & 0.40                                  & 0.79                                  & 0.53                                  & 0.53                                  & 0.53                                  & 0.73                                  & 0.70                                  & 0.66                                  & 0.66                                  & 0.70                                  & 0.54                                  \\
                                                                                   & \textbf{Autoformer}                              & 0.72                                  & 0.56                                  & 0.56                                  & 0.54                                  & 0.62                                  & 0.47                                  & 0.44                                  & 0.44                                  & 0.79                                  & 0.52                                  & 0.51                                  & 0.51                                  & 0.70                                  & 0.64                                  & 0.64                                  & 0.61                                  & 0.71                                  & 0.53                                  \\
                                                                                   & \textbf{Crossformer}                             & 0.66                                  & 0.58                                  & 0.51                                  & 0.53                                  & 0.65                                  & 0.55                                  & 0.48                                  & 0.50                                  & 0.79                                  & 0.50                                  & 0.51                                  & 0.50                                  & 0.72                                  & 0.71                                  & 0.58                                  & 0.58                                  & 0.71                                  & 0.53                                  \\
                                                                                   & \textbf{FEDformer}                               & 0.67                                  & 0.57                                  & 0.50                                  & 0.51                                  & 0.65                                  & 0.53                                  & 0.47                                  & 0.49                                  & 0.76                                  & 0.57                                  & 0.58                                  & 0.57                                  & 0.68                                  & 0.48                                  & 0.54                                  & 0.48                                  & 0.69                                  & 0.51                                  \\
                                                                                   & \textbf{Informer}                                & 0.67                                  & 0.59                                  & 0.51                                  & 0.52                                  & 0.67                                  & 0.59                                  & 0.51                                  & 0.52                                  & 0.82                                  & 0.57                                  & 0.55                                  & 0.55                                  & \underline{ 0.77}                            & 0.74                                  & 0.69                                  & 0.71                                  & 0.73                                  & 0.58                                  \\
                                                                                   & \textbf{Reformer}                                & 0.69                                  & 0.56                                  & 0.53                                  & 0.54                                  & 0.65                                  & 0.53                                  & 0.48                                  & 0.49                                  & \underline{ 0.84}                            & 0.52                                  & 0.50                                  & 0.48                                  & 0.74                                  & 0.75                                  & 0.61                                  & 0.66                                  & 0.73                                  & 0.54                                  \\
                                                                                   & \textbf{iTransformer}                            & 0.56                                  & 0.42                                  & 0.36                                  & 0.37                                  & 0.54                                  & 0.39                                  & 0.31                                  & 0.29                                  & 0.80                                  & 0.50                                  & 0.50                                  & 0.49                                  & 0.73                                  & 0.75                                  & 0.59                                  & 0.61                                  & 0.66                                  & 0.44                                  \\
\multirow{-13}{*}{\textbf{Time}}                                                   & \textbf{PatchTST}                                & \underline{ 0.78}                            & \underline{ 0.76}                            & 0.62                                  & 0.62                                  & \underline{ 0.74}                            & \underline{ 0.69}                            & 0.59                                  & 0.62                                  & 0.73                                  & 0.54                                  & 0.55                                  & 0.54                                  & 0.70                                  & 0.65                                  & 0.59                                  & 0.57                                  & 0.74                                  & 0.59                                  \\ \midrule
\textbf{Time}                                                                      & \cellcolor[HTML]{E8E8E8}\textbf{GPT4TS}          & \cellcolor[HTML]{E8E8E8}0.71          & \cellcolor[HTML]{E8E8E8}0.58          & \cellcolor[HTML]{E8E8E8}0.52          & \cellcolor[HTML]{E8E8E8}0.53          & \cellcolor[HTML]{E8E8E8}0.59          & \cellcolor[HTML]{E8E8E8}0.46          & \cellcolor[HTML]{E8E8E8}0.45          & \cellcolor[HTML]{E8E8E8}0.45          & \cellcolor[HTML]{E8E8E8}0.78          & \cellcolor[HTML]{E8E8E8}0.48          & \cellcolor[HTML]{E8E8E8}0.48          & \cellcolor[HTML]{E8E8E8}0.48          & \cellcolor[HTML]{E8E8E8}0.71          & \cellcolor[HTML]{E8E8E8}0.73          & \cellcolor[HTML]{E8E8E8}0.60          & \cellcolor[HTML]{E8E8E8}0.64          & \cellcolor[HTML]{E8E8E8}0.70          & \cellcolor[HTML]{E8E8E8}0.53          \\ \midrule
                                                                                   & \cellcolor[HTML]{E8E8E8}\textbf{GPT2}            & \cellcolor[HTML]{E8E8E8}0.72          & \cellcolor[HTML]{E8E8E8}0.65          & \cellcolor[HTML]{E8E8E8}0.56          & \cellcolor[HTML]{E8E8E8}0.58          & \cellcolor[HTML]{E8E8E8}0.73          & \cellcolor[HTML]{E8E8E8}0.65          & \cellcolor[HTML]{E8E8E8}0.61          & \cellcolor[HTML]{E8E8E8}0.62          & \cellcolor[HTML]{E8E8E8}0.72          & \cellcolor[HTML]{E8E8E8}0.49          & \cellcolor[HTML]{E8E8E8}0.49          & \cellcolor[HTML]{E8E8E8}0.50          & \cellcolor[HTML]{E8E8E8}0.64          & \cellcolor[HTML]{E8E8E8}0.69          & \cellcolor[HTML]{E8E8E8}0.53          & \cellcolor[HTML]{E8E8E8}0.58          & \cellcolor[HTML]{E8E8E8}0.70          & \cellcolor[HTML]{E8E8E8}0.57          \\
                                                                                   & \cellcolor[HTML]{E8E8E8}\textbf{BERT}            & \cellcolor[HTML]{E8E8E8}0.70          & \cellcolor[HTML]{E8E8E8}0.64          & \cellcolor[HTML]{E8E8E8}0.51          & \cellcolor[HTML]{E8E8E8}0.53          & \cellcolor[HTML]{E8E8E8}0.73          & \cellcolor[HTML]{E8E8E8}0.65          & \cellcolor[HTML]{E8E8E8}0.59          & \cellcolor[HTML]{E8E8E8}0.62          & \cellcolor[HTML]{E8E8E8}0.72          & \cellcolor[HTML]{E8E8E8}0.49          & \cellcolor[HTML]{E8E8E8}0.49          & \cellcolor[HTML]{E8E8E8}0.49          & \cellcolor[HTML]{E8E8E8}0.59          & \cellcolor[HTML]{E8E8E8}0.45          & \cellcolor[HTML]{E8E8E8}0.39          & \cellcolor[HTML]{E8E8E8}0.40          & \cellcolor[HTML]{E8E8E8}0.69          & \cellcolor[HTML]{E8E8E8}0.51          \\
                                                                                   & \cellcolor[HTML]{E8E8E8}\textbf{Llama 3}         & \cellcolor[HTML]{E8E8E8}0.73          & \cellcolor[HTML]{E8E8E8}0.60          & \cellcolor[HTML]{E8E8E8}0.60          & \cellcolor[HTML]{E8E8E8}0.60          & \cellcolor[HTML]{E8E8E8}\underline{ 0.74}    & \cellcolor[HTML]{E8E8E8}\underline{ 0.69}    & \cellcolor[HTML]{E8E8E8}0.56          & \cellcolor[HTML]{E8E8E8}0.65          & \cellcolor[HTML]{E8E8E8}0.72          & \cellcolor[HTML]{E8E8E8}0.53          & \cellcolor[HTML]{E8E8E8}0.53          & \cellcolor[HTML]{E8E8E8}0.55          & \cellcolor[HTML]{E8E8E8}0.66          & \cellcolor[HTML]{E8E8E8}0.62          & \cellcolor[HTML]{E8E8E8}0.47          & \cellcolor[HTML]{E8E8E8}0.47          & \cellcolor[HTML]{E8E8E8}0.71          & \cellcolor[HTML]{E8E8E8}0.57          \\
\multirow{-4}{*}{\textbf{Text}}                                                    & \cellcolor[HTML]{E8E8E8}\textbf{ClinicalBERT}    & \cellcolor[HTML]{E8E8E8}0.73          & \cellcolor[HTML]{E8E8E8}0.57          & \cellcolor[HTML]{E8E8E8}0.54          & \cellcolor[HTML]{E8E8E8}0.53          & \cellcolor[HTML]{E8E8E8}\underline{ 0.74}    & \cellcolor[HTML]{E8E8E8}0.63          & \cellcolor[HTML]{E8E8E8}0.58          & \cellcolor[HTML]{E8E8E8}\underline{ 0.66}    & \cellcolor[HTML]{E8E8E8}0.72          & \cellcolor[HTML]{E8E8E8}0.55          & \cellcolor[HTML]{E8E8E8}0.63          & \cellcolor[HTML]{E8E8E8}0.56          & \cellcolor[HTML]{E8E8E8}0.67          & \cellcolor[HTML]{E8E8E8}0.36          & \cellcolor[HTML]{E8E8E8}0.64          & \cellcolor[HTML]{E8E8E8}0.43          & \cellcolor[HTML]{E8E8E8}0.72          & \cellcolor[HTML]{E8E8E8}0.55          \\ \midrule
                                                                                   & \cellcolor[HTML]{E8E8E8}\textbf{TimeLLM}         & \cellcolor[HTML]{E8E8E8}0.69          & \cellcolor[HTML]{E8E8E8}0.60          & \cellcolor[HTML]{E8E8E8}0.48          & \cellcolor[HTML]{E8E8E8}0.47          & \cellcolor[HTML]{E8E8E8}0.67          & \cellcolor[HTML]{E8E8E8}0.59          & \cellcolor[HTML]{E8E8E8}0.46          & \cellcolor[HTML]{E8E8E8}0.48          & \cellcolor[HTML]{E8E8E8}0.75          & \cellcolor[HTML]{E8E8E8}0.51          & \cellcolor[HTML]{E8E8E8}0.51          & \cellcolor[HTML]{E8E8E8}0.51          & \cellcolor[HTML]{E8E8E8}0.69          & \cellcolor[HTML]{E8E8E8}0.70          & \cellcolor[HTML]{E8E8E8}0.50          & \cellcolor[HTML]{E8E8E8}0.47          & \cellcolor[HTML]{E8E8E8}0.70          & \cellcolor[HTML]{E8E8E8}0.48          \\
                                                                                   & \cellcolor[HTML]{E8E8E8}\textbf{UniTime}         & \cellcolor[HTML]{E8E8E8}0.67          & \cellcolor[HTML]{E8E8E8}0.33          & \cellcolor[HTML]{E8E8E8}0.42          & \cellcolor[HTML]{E8E8E8}0.37          & \cellcolor[HTML]{E8E8E8}0.64          & \cellcolor[HTML]{E8E8E8}0.54          & \cellcolor[HTML]{E8E8E8}0.43          & \cellcolor[HTML]{E8E8E8}0.44          & \cellcolor[HTML]{E8E8E8}0.79          & \cellcolor[HTML]{E8E8E8}0.54          & \cellcolor[HTML]{E8E8E8}0.53          & \cellcolor[HTML]{E8E8E8}0.53          & \cellcolor[HTML]{E8E8E8}\underline{ 0.77}    & \cellcolor[HTML]{E8E8E8}\underline{ 0.78}    & \cellcolor[HTML]{E8E8E8}\underline{ 0.71}    & \cellcolor[HTML]{E8E8E8}0.71          & \cellcolor[HTML]{E8E8E8}0.72          & \cellcolor[HTML]{E8E8E8}0.51          \\
                                                                                   & \cellcolor[HTML]{E8E8E8}\textbf{GPT4MTS}         & \cellcolor[HTML]{E8E8E8}0.72          & \cellcolor[HTML]{E8E8E8}0.59          & \cellcolor[HTML]{E8E8E8}0.60          & \cellcolor[HTML]{E8E8E8}0.59          & \cellcolor[HTML]{E8E8E8}0.65          & \cellcolor[HTML]{E8E8E8}0.48          & \cellcolor[HTML]{E8E8E8}0.50          & \cellcolor[HTML]{E8E8E8}0.48          & \cellcolor[HTML]{E8E8E8}0.82          & \cellcolor[HTML]{E8E8E8}0.64          & \cellcolor[HTML]{E8E8E8}0.63          & \cellcolor[HTML]{E8E8E8}\underline{ 0.63}    & \cellcolor[HTML]{E8E8E8}0.70          & \cellcolor[HTML]{E8E8E8}0.72          & \cellcolor[HTML]{E8E8E8}0.60          & \cellcolor[HTML]{E8E8E8}0.53          & \cellcolor[HTML]{E8E8E8}0.72          & \cellcolor[HTML]{E8E8E8}0.56          \\
                                                                                   & \cellcolor[HTML]{E8E8E8}\textbf{MedTsLLM}        & \cellcolor[HTML]{E8E8E8}0.74          & \cellcolor[HTML]{E8E8E8}0.64          & \cellcolor[HTML]{E8E8E8}0.63          & \cellcolor[HTML]{E8E8E8}\underline{ 0.64}    & \cellcolor[HTML]{E8E8E8}0.68          & \cellcolor[HTML]{E8E8E8}0.64          & \cellcolor[HTML]{E8E8E8}0.60          & \cellcolor[HTML]{E8E8E8}0.56          & \cellcolor[HTML]{E8E8E8}0.81          & \cellcolor[HTML]{E8E8E8}0.66          & \cellcolor[HTML]{E8E8E8}\underline{ 0.64}    & \cellcolor[HTML]{E8E8E8}\underline{ 0.64}    & \cellcolor[HTML]{E8E8E8}0.72          & \cellcolor[HTML]{E8E8E8}0.74          & \cellcolor[HTML]{E8E8E8}0.59          & \cellcolor[HTML]{E8E8E8}0.60          & \cellcolor[HTML]{E8E8E8}0.74          & \cellcolor[HTML]{E8E8E8}\underline{ 0.61}    \\ \cmidrule{2-20}
                                                                                   & \cellcolor[HTML]{E8E8E8}\textbf{MedualTime (Time)} & \cellcolor[HTML]{E8E8E8}0.73          & \cellcolor[HTML]{E8E8E8}0.65          & \cellcolor[HTML]{E8E8E8}\underline{ 0.66}    & \cellcolor[HTML]{E8E8E8}0.62          & \cellcolor[HTML]{E8E8E8}0.71          & \cellcolor[HTML]{E8E8E8}0.62          & \cellcolor[HTML]{E8E8E8}\underline{ 0.64}    & \cellcolor[HTML]{E8E8E8}0.59          & \cellcolor[HTML]{E8E8E8}0.83          & \cellcolor[HTML]{E8E8E8}\underline{ 0.68}    & \cellcolor[HTML]{E8E8E8}0.63          & \cellcolor[HTML]{E8E8E8}0.60          & \cellcolor[HTML]{E8E8E8}0.74          & \cellcolor[HTML]{E8E8E8}0.75          & \cellcolor[HTML]{E8E8E8}0.61          & \cellcolor[HTML]{E8E8E8}0.62          & \cellcolor[HTML]{E8E8E8}\underline{ 0.75}    & \cellcolor[HTML]{E8E8E8}\underline{ 0.61}    \\
                                                                                   & \cellcolor[HTML]{E8E8E8}\textbf{MedualTime (Text)} & \cellcolor[HTML]{E8E8E8}\underline{ 0.79}    & \cellcolor[HTML]{E8E8E8}\underline{ 0.73}    & \cellcolor[HTML]{E8E8E8}\underline{ 0.71}    & \cellcolor[HTML]{E8E8E8}\underline{ 0.70}    & \cellcolor[HTML]{E8E8E8}\underline{ 0.75}    & \cellcolor[HTML]{E8E8E8}\underline{ 0.66}    & \cellcolor[HTML]{E8E8E8}\underline{ 0.67}    & \cellcolor[HTML]{E8E8E8}\underline{ 0.66}    & \cellcolor[HTML]{E8E8E8}0.82          & \cellcolor[HTML]{E8E8E8}\underline{ 0.67}    & \cellcolor[HTML]{E8E8E8}\underline{ 0.64}    & \cellcolor[HTML]{E8E8E8}\underline{ 0.63}    & \cellcolor[HTML]{E8E8E8}\underline{ 0.75}    & \cellcolor[HTML]{E8E8E8}0.76          & \cellcolor[HTML]{E8E8E8}0.68          & \cellcolor[HTML]{E8E8E8}0.65          & \cellcolor[HTML]{E8E8E8}\underline{ 0.78}    & \cellcolor[HTML]{E8E8E8}\underline{ 0.66}    \\
\multirow{-7}{*}{\textbf{\begin{tabular}[c]{@{}c@{}}Time\\ +\\ Text\end{tabular}}} & \cellcolor[HTML]{E8E8E8}\textbf{MedualTime}        & \cellcolor[HTML]{E8E8E8}\textbf{0.83} & \cellcolor[HTML]{E8E8E8}\textbf{0.78} & \cellcolor[HTML]{E8E8E8}\textbf{0.75} & \cellcolor[HTML]{E8E8E8}\textbf{0.76} & \cellcolor[HTML]{E8E8E8}\textbf{0.81} & \cellcolor[HTML]{E8E8E8}\textbf{0.75} & \cellcolor[HTML]{E8E8E8}\textbf{0.75} & \cellcolor[HTML]{E8E8E8}\textbf{0.75} & \cellcolor[HTML]{E8E8E8}\textbf{0.87} & \cellcolor[HTML]{E8E8E8}\textbf{0.75} & \cellcolor[HTML]{E8E8E8}\textbf{0.65} & \cellcolor[HTML]{E8E8E8}\textbf{0.68} & \cellcolor[HTML]{E8E8E8}\textbf{0.79} & \cellcolor[HTML]{E8E8E8}\textbf{0.82} & \cellcolor[HTML]{E8E8E8}\textbf{0.74} & \cellcolor[HTML]{E8E8E8}\textbf{0.75} & \cellcolor[HTML]{E8E8E8}\textbf{0.83} & \cellcolor[HTML]{E8E8E8}\textbf{0.73}

                                                                                    \\ \bottomrule

\end{tabular}
\end{threeparttable}
}
\caption{\textbf{Supervised Learning}. MedualTime achieves an average improvement of \textbf{8\%} in Acc. and \textbf{12\%} in F1 compared to baselines. The best results are in \textbf{bold} while the second and third best are in \underline{underlined}. "Acc.", "Pre.", and "Rec." represent accuracy, precision and recall respectively. All LM-based models are highlighted in grey.}
\label{tab:sup}
\end{table*}

\subsection{Experimental Setup}

\textbf{Datasets} Since our approach focuses on medical tasks involving coupled time-series and text modalities, where each modality provides valuable information for decision-making, we conduct experiments using publicly available datasets that meet the necessary criteria above: (1) PTB-XL \footnote{\url{https://physionet.org/content/ptb-xl/1.0.3/}} \cite{wagner2020ptb}: This electrocardiogram (ECG) corpus consists of 12-lead ECG signals, which capture the electrical activity of the heart, along with clinical reports describing signal characteristics without diagnostic labels. PTB-XL provides two label sets: a coarse-grained label set for disease detection (four classes) and a fine-grained label set for specific disease classification (five classes).
(2) TUSZ v1.5.2 \footnote{\url{https://isip.piconepress.com/projects/nedc/html/tuh_eeg/}} \cite{shah2018temple}: The Temple University Seizure Corpus (TUSZ) is a large-scale corpus of EEG (electroencephalogram) signals that record the electrical activity of the brain. It includes 19-channel EEG recordings and the clinical history for each patient session. Similar to PTB-XL, TUSZ offers two label sets: a coarse-grained label set for distinguishing seizure and non-seizure EEG signals (two classes), and a fine-grained label set for seizure type classification (four classes). More details about the datasets, including the label sets, data splits, and preprocessing steps, are provided in \textbf{Appendix 1.1}.

\textbf{Baselines} Representative baselines are selected to ensure sufficient experiments. 
(1) \textbf{Unimodal LM-free methods}: MLP-based models (LightTS \cite{zhang2022lightts}, DLinear \cite{zeng2023dlinear}); RNN-based models (LSTM \cite{hochreiter1997lstm}); CNN-based models (TimesNet \cite{wu2022timesnet}, TS2Vec \cite{yue2022ts2vec}, TS-CoT \cite{zhang2023co}); Transformer-based models (Pyraformer \cite{liu2021pyraformer}, ETSformer \cite{woo2022etsformer}, Autoformer \cite{wu2021autoformer}, Crossformer \cite{zhang2022crossformer}, FEDformer \cite{zhou2022fedformer}, Informer \cite{zhou2021informer},  Reformer \cite{kitaev2020reformer}, iTransformer \cite{liu2023itransformer}, PatchTST \cite{nie2022patchtst}, TS-TCC \cite{tstcc}). 
(2) \textbf{Unimodal LM-based methods}: BERT \cite{devlin2018bert}, GPT-2 \cite{radford2019gpt2}, Llama 3, ClinicalBERT \cite{wang2023optimized}, GPT4TS \cite{zhou2024one}. 
(3) \textbf{Multimodal LM-based methods}: TimeLLM \cite{jin2023timellm}, UniTime \cite{liu2023unitime}, GPT4MTS \cite{jia2024gpt4mts}, MedTsLLM \cite{chan2024medtsllm} for supervised learning; METS \cite{li2024frozen}, MERL \cite{liuzero} for unsupervised learning.
(4) \textbf{MedualTime variants}: \textit{MedualTime (Time)} for temporal-primary multimodal adapter, \textit{MedualTime (Text)} for textual-primary multimodal adapter. Note that for GPT-2, BERT, Llama 3, and ClinicalBERT,  we use textual embeddings generated by them and then train a linear classifier from scratch for the downstream task. More introductions about baselines are in \textbf{Appendix 1.4}.



\textbf{Implementations} Inspired by \cite{liu2023unitime,jia2024gpt4mts,zhou2024one}, which have validated GPT-2's capability in handling time-series data, MedualTime adopts a frozen GPT-2 as the backbone. In the textual-primary multimodal adapter, the tokenizer is derived from GPT-2. To avoid heavy computational costs, we use a lightweight CNN-based model as the temporal encoder, consisting of three convolutional blocks, each with three CNN layers. This encoder is trained from scratch to adapt to the specific tasks. In the temporal-primary multimodal adapter, following METS \cite{li2024frozen}, a frozen ClinicalBERT \cite{wang2023optimized} pre-trained on medical data as the textual encoder. 
All hidden dimensions are set to 768 to match the backbone (i.e., GPT-2). The time series patch size and stride are both set to 25. Adam is adopted as the optimizer \cite{kingma2014adam}. All experiments are implemented by PyTorch Framework with NVIDIA A6000 (48G) GPU.

\subsection{Supervised Learning}
\label{subsec:sup}

A linear classifier is added as the output layer of MedualTime under supervised learning. As shown in Table \ref{tab:sup}, \textbf{(1)} Time-only models perform better than text-only models, achieving second best in most experiments.
\textbf{(2)} Among multimodal approaches based on LMs, UniTime and GPT4MTS exhibit similar performance, outperforming TimeLLM by a 2\% accuracy improvement. This performance gap may be due to the differences in their fine-tuning strategies. While TimeLLM relies on a frozen LLM, UniTime and GPT4MTS employ parameter-efficient fine-tuning techniques. MedTsLLM outperforms GPT4MTS, perhaps due to its tailored design for medical data.
\textbf{(3)} MedualTime significantly outperforms these LM-based multimodal methods by a 9\% accuracy improvement. This discrepancy likely arises from their temporal-primary paradigm, which overlooks critical information embedded in the text modality. In contrast, MedualTime utilizes a textual-temporal perspective, allowing for a more comprehensive understanding of the multimodal interactions.
\textbf{(4)} MedualTime achieves the best performance, improving accuracy by \textbf{8\%} and F1 by \textbf{12\%} on average.

\begin{figure}[htb]
\vspace{-2mm}
  \begin{center}
    \includegraphics[width=0.3\textwidth]{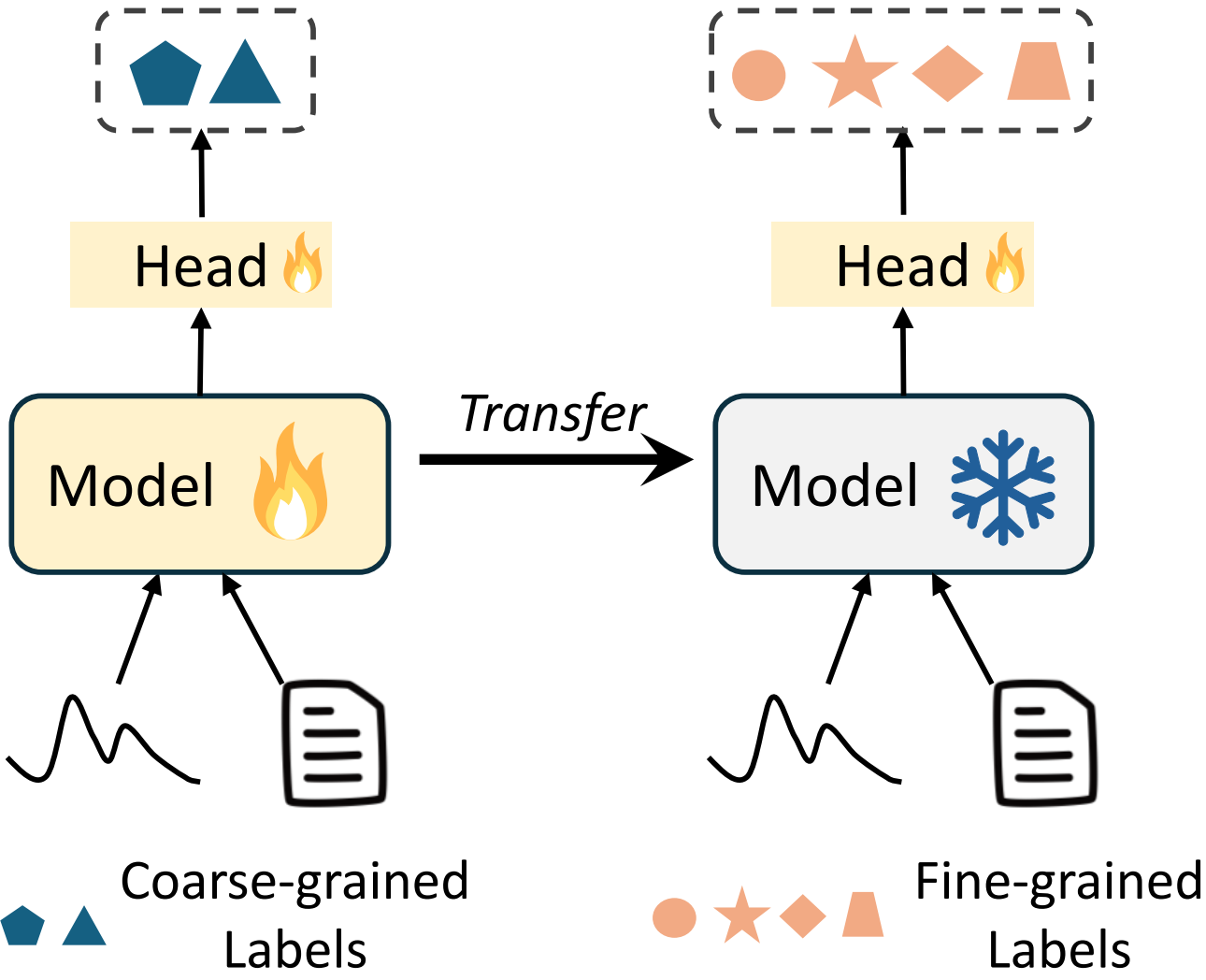} 
  \end{center}
  \vspace{-3mm}
  \caption{Illustration for \textbf{Few-shot Label Transfer}.}

  
  \label{fig:lab}
  \vspace{-3mm}
\end{figure}
\begin{figure}[htb]
\centering

    \includegraphics[width=0.49\textwidth]{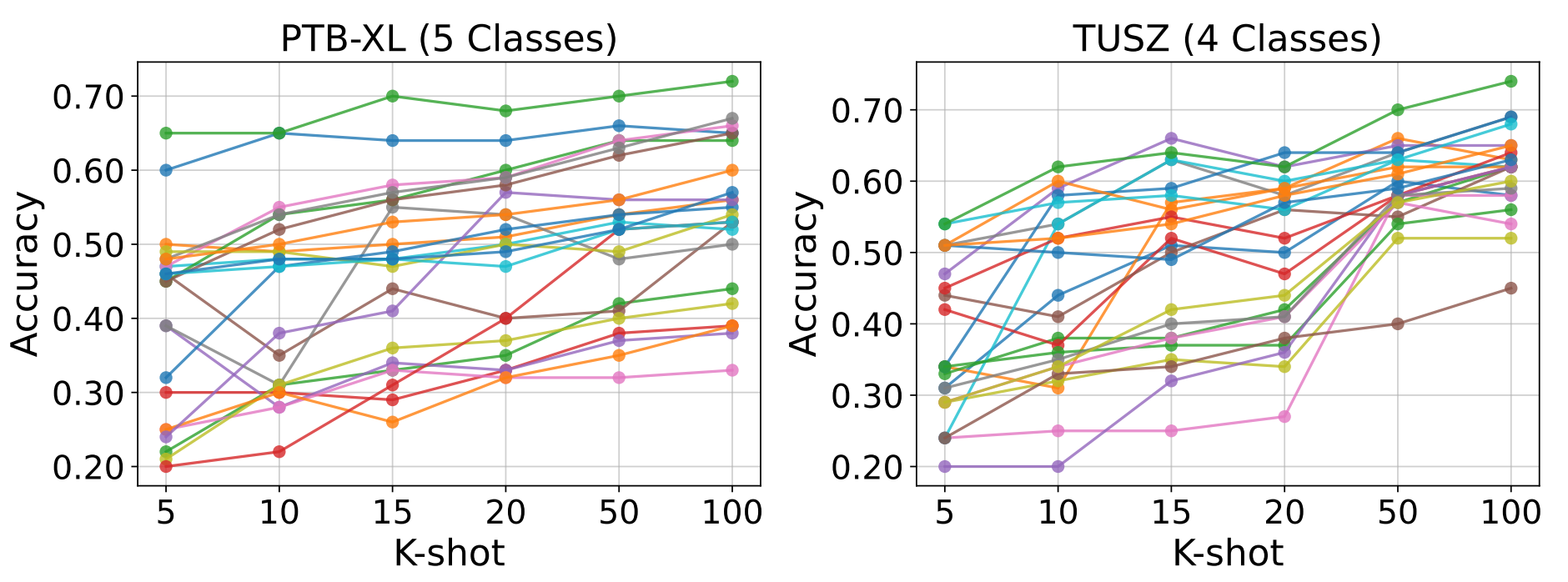}
      
    \vspace{-3mm} 
    \caption{Few-shot experiment results on accuracy. The green line is MedualTime.}
\label{fig:few}
\vspace{-3mm} 
\end{figure}

\subsection{Few-shot Learning for Label Transfer}
\label{subsec:few}


To evaluate the transferability of representations in a supervised setting, we propose a \textbf{Few-shot Label Transfer} setting for in-dataset transfer between coarse-grained and fine-grained label sets (Figure \ref{fig:lab}). It is common that in real-world applications coarse-grained labels, such as disease presence, are easier to obtain, while fine-grained labels, like specific disease types, are expensive to obtain. In this setting, we pre-train the model on coarse-grained labels and fine-tune it on limited fine-grained labels by freezing the pre-trained parameters and training an additional classifier for few-shot learning. Specifically, we train the model on PTB-XL with four classes and fine-tune on five classes, and train on TUSZ with two classes and fine-tune on four classes. We conduct \{5, 10, 15, 20, 50, 100\}-shot experiments. Figure \ref{fig:few} shows the results on accuracy and more results are in \textbf{Appendix 1.3}. Generally speaking, all models'  accuracy generally shows a continuous growth trend as the few-shot (K) increases. MedualTime consistently outperforms baselines on both datasets and achieves strong classification performance even with limited training samples, demonstrating its effectiveness and robust transferability powered by language models and multimodal inputs.

\begin{table}[htb] 

\centering 
    
    \resizebox{\columnwidth}{!}{ 
    
        \begin{threeparttable} 
        
            

\begin{tabular}{c|c|cc|cc|cc|cc|cc}
\toprule

\multicolumn{1}{c|}{} &
  \multicolumn{1}{c|}{} &
  \multicolumn{4}{c|}{\textbf{PTB-XL}} &
  \multicolumn{4}{c|}{\textbf{TUSZ}} &
  \multicolumn{2}{c}{} \\ \cmidrule{3-10}
\multicolumn{1}{c|}{} &
  \multicolumn{1}{c|}{} &
  \multicolumn{2}{c}{\textbf{4 Classes}} &
  \multicolumn{2}{c|}{\textbf{5 Classes}} &
  \multicolumn{2}{c}{\textbf{2 Classes}} &
  \multicolumn{2}{c|}{\textbf{4 Classes}} &
  \multicolumn{2}{c}{\multirow{-2}{*}{\textbf{Average}}} \\
\multicolumn{1}{c|}{\multirow{-3}{*}{\textbf{Modality}}} &
  \multicolumn{1}{c|}{\multirow{-3}{*}{\textbf{Model}}} &
  \multicolumn{1}{c}{\textbf{Acc.}} &
  \multicolumn{1}{c}{\textbf{F1}} &
  \multicolumn{1}{c}{\textbf{Acc.}} &
  \multicolumn{1}{c|}{\textbf{F1}} &
  \multicolumn{1}{c}{\textbf{Acc.}} &
  \multicolumn{1}{c}{\textbf{F1}} &
  \multicolumn{1}{c}{\textbf{Acc.}} &
  \multicolumn{1}{c|}{\textbf{F1}} &
  \multicolumn{1}{c}{\textbf{Acc.}} &
  \multicolumn{1}{c}{\textbf{F1}} \\ \midrule
                                                                                   & \textbf{TSTCC}                                   & 0.68                                  & 0.54                                  & 0.65                                  & 0.50                                  & \underline{ 0.74}                            & 0.48                                  & 0.67                                  & 0.45                                  & 0.69                                  & 0.49                                  \\
                                                                                   & \textbf{TS2Vec}                                  & 0.61                                  & 0.43                                  & 0.61                                  & 0.49                                  & 0.70                                  & 0.49                                  & \underline{ 0.70}                            & 0.53                                  & 0.66                                  & 0.48                                  \\
                                                                                   & \textbf{TSCoT}                                   & 0.73                                  & \underline{ 0.60}                            & \underline{ 0.75}                            & 0.63                                  & 0.67                                  & 0.53                                  & 0.69                                  & \underline{ 0.60}                            & 0.71                                  & \underline{ 0.59}                            \\
\multirow{-4}{*}{\textbf{Time}}                                                    & \textbf{PatchTST}                                & 0.60                                  & 0.35                                  & 0.55                                  & 0.30                                  & \underline{ 0.73}                            & 0.50                                  & 0.67                                  & 0.45                                  & 0.64                                  & 0.40                                  \\ \midrule
                                                                                   & \cellcolor[HTML]{D9D9D9}\textbf{GPT2}            & \cellcolor[HTML]{D9D9D9}0.72          & \cellcolor[HTML]{D9D9D9}0.58          & \cellcolor[HTML]{D9D9D9}0.73          & \cellcolor[HTML]{D9D9D9}0.62          & \cellcolor[HTML]{D9D9D9}0.72          & \cellcolor[HTML]{D9D9D9}0.50          & \cellcolor[HTML]{D9D9D9}0.64          & \cellcolor[HTML]{D9D9D9}0.39          & \cellcolor[HTML]{D9D9D9}0.70          & \cellcolor[HTML]{D9D9D9}0.52          \\
                                                                                   & \cellcolor[HTML]{D9D9D9}\textbf{Bert}            & \cellcolor[HTML]{D9D9D9}0.70          & \cellcolor[HTML]{D9D9D9}0.53          & \cellcolor[HTML]{D9D9D9}0.73          & \cellcolor[HTML]{D9D9D9}0.62          & \cellcolor[HTML]{D9D9D9}0.72          & \cellcolor[HTML]{D9D9D9}0.49          & \cellcolor[HTML]{D9D9D9}0.59          & \cellcolor[HTML]{D9D9D9}0.40          & \cellcolor[HTML]{D9D9D9}0.69          & \cellcolor[HTML]{D9D9D9}0.51          \\
                                                                                   & \cellcolor[HTML]{D9D9D9}\textbf{Llama 3}         & \cellcolor[HTML]{D9D9D9}0.73          & \cellcolor[HTML]{D9D9D9}\underline{ 0.60}    & \cellcolor[HTML]{D9D9D9}0.74          & \cellcolor[HTML]{D9D9D9}0.55          & \cellcolor[HTML]{D9D9D9}0.72          & \cellcolor[HTML]{D9D9D9}\underline{ 0.63}    & \cellcolor[HTML]{D9D9D9}0.66          & \cellcolor[HTML]{D9D9D9}0.47          & \cellcolor[HTML]{D9D9D9}0.71          & \cellcolor[HTML]{D9D9D9}0.56          \\
\multirow{-4}{*}{\textbf{Text}}                                                    & \cellcolor[HTML]{D9D9D9}\textbf{ClinicalBERT}    & \cellcolor[HTML]{D9D9D9}0.73          & \cellcolor[HTML]{D9D9D9}0.53          & \cellcolor[HTML]{D9D9D9}0.74          & \cellcolor[HTML]{D9D9D9}0.55          & \cellcolor[HTML]{D9D9D9}0.72          & \cellcolor[HTML]{D9D9D9}\underline{ 0.66}    & \cellcolor[HTML]{D9D9D9}0.67          & \cellcolor[HTML]{D9D9D9}0.43          & \cellcolor[HTML]{D9D9D9}\underline{ 0.72}    & \cellcolor[HTML]{D9D9D9}0.54          \\ \midrule
                                                                                   & \cellcolor[HTML]{D9D9D9}\textbf{METS}            & \cellcolor[HTML]{D9D9D9}0.74          & \cellcolor[HTML]{D9D9D9}0.58          & \cellcolor[HTML]{D9D9D9}0.71          & \cellcolor[HTML]{D9D9D9}0.60          & \cellcolor[HTML]{D9D9D9}0.65          & \cellcolor[HTML]{D9D9D9}0.53          & \cellcolor[HTML]{D9D9D9}0.57          & \cellcolor[HTML]{D9D9D9}0.20          & \cellcolor[HTML]{D9D9D9}0.67          & \cellcolor[HTML]{D9D9D9}0.48          \\
                                                                                   & \cellcolor[HTML]{D9D9D9}\textbf{MERL}            & \cellcolor[HTML]{D9D9D9}\underline{ 0.75}    & \cellcolor[HTML]{D9D9D9}0.58          & \cellcolor[HTML]{D9D9D9}\underline{ 0.75}    & \cellcolor[HTML]{D9D9D9}\underline{ 0.66}    & \cellcolor[HTML]{D9D9D9}0.70          & \cellcolor[HTML]{D9D9D9}0.57          & \cellcolor[HTML]{D9D9D9}\underline{ 0.70}    & \cellcolor[HTML]{D9D9D9}0.50          & \cellcolor[HTML]{D9D9D9}\underline{ 0.73}    & \cellcolor[HTML]{D9D9D9}\underline{ 0.58}    \\ \cmidrule{2-12}
                                                                                   & \cellcolor[HTML]{D9D9D9}\textbf{MedualTime (Time)} & \cellcolor[HTML]{D9D9D9}0.71          & \cellcolor[HTML]{D9D9D9}0.47          & \cellcolor[HTML]{D9D9D9}0.65          & \cellcolor[HTML]{D9D9D9}0.47          & \cellcolor[HTML]{D9D9D9}0.69          & \cellcolor[HTML]{D9D9D9}0.52          & \cellcolor[HTML]{D9D9D9}0.69          & \cellcolor[HTML]{D9D9D9}0.44          & \cellcolor[HTML]{D9D9D9}0.69          & \cellcolor[HTML]{D9D9D9}0.48          \\
                                                                                   & \cellcolor[HTML]{D9D9D9}\textbf{MedualTime (Text)} & \cellcolor[HTML]{D9D9D9}0.73          & \cellcolor[HTML]{D9D9D9}0.58          & \cellcolor[HTML]{D9D9D9}0.71          & \cellcolor[HTML]{D9D9D9}0.62          & \cellcolor[HTML]{D9D9D9}0.71          & \cellcolor[HTML]{D9D9D9}0.52          & \cellcolor[HTML]{D9D9D9}\underline{ 0.71}    & \cellcolor[HTML]{D9D9D9}\underline{ 0.62}    & \cellcolor[HTML]{D9D9D9}\underline{ 0.72}    & \cellcolor[HTML]{D9D9D9}\underline{ 0.59}    \\
\multirow{-5}{*}{\textbf{\begin{tabular}[c]{@{}c@{}}Time\\ +\\ Text\end{tabular}}} & \cellcolor[HTML]{D9D9D9}\textbf{MedualTime}        & \cellcolor[HTML]{D9D9D9}\textbf{0.77} & \cellcolor[HTML]{D9D9D9}\textbf{0.65} & \cellcolor[HTML]{D9D9D9}\textbf{0.77} & \cellcolor[HTML]{D9D9D9}\textbf{0.58} & \cellcolor[HTML]{D9D9D9}\textbf{0.76} & \cellcolor[HTML]{D9D9D9}\textbf{0.57} & \cellcolor[HTML]{D9D9D9}\textbf{0.74} & \cellcolor[HTML]{D9D9D9}\textbf{0.62} & \cellcolor[HTML]{D9D9D9}\textbf{0.76} & \cellcolor[HTML]{D9D9D9}\textbf{0.61}

            \\ 
            \bottomrule 
            \end{tabular} 
            
        
        \end{threeparttable} 
        
        } 

\caption{\textbf{Unsupervised Learning}. 100\% of the labeled training data are used to train a linear classifier.} 
\label{tab:uns} 
\end{table} 

\subsection{Unsupervised Learning}
\label{subsec:uns}

We follow the experimental setting in previous works \cite{zhang2023co} to assess our model's ability to generate general representations without ground truth supervision. First, we conduct unsupervised experiments. Then, using the unsupervised embeddings obtained, we train a linear classifier with varying proportions (from 10\% to 100\%) of labeled training data. Table \ref{tab:uns} shows the results for the 100\% labeled training data proportion. 

As shown in Table \ref{tab:uns}, text-only models generally outperform time-only models, highlighting the importance of text data in the unsupervised setting. While the multimodal model MERL outperforms the best time-only model TSCoT, METS lags behind, suggesting that multimodal does not always outperform single modality. The effectiveness of multimodal fusion is crucial. MedualTime surpasses MERL in most experiments, emphasizing the advantages of our complementary textual-temporal multimodal design. Overall, MedualTime achieves an average accuracy improvement of 3\% and F1 improvement of 2\%, consistently outperforming other baselines.

\subsection{More Explorations}
\label{subsec:exp}
\textbf{(a) Ablation Study} We ablate MedualTime into MedualTime (Time) and MedualTime (Text) under both supervised and unsupervised settings, as shown in Tables \ref{tab:sup} and \ref{tab:uns}. Generally, MedualTime (Text) achieves better performance than MedualTime (Time) in both settings, suggesting that the backbone language model demonstrates a stronger understanding of text compared to time series. Overall, MedualTime consistently outperforms the single-adapter variants, highlighting the complementary nature of textual and temporal information.

\begin{table}[htb] 

\centering 
    
    \resizebox{\columnwidth}{!}{ 
    
        \begin{threeparttable} 
        
            

\begin{tabular}{c|cc|cc|cc|cc|cc}
\toprule

\multicolumn{1}{c|}{\multirow{3}{*}{\textbf{Model}}} &
  \multicolumn{4}{c|}{\textbf{PTB-XL}} &
  \multicolumn{4}{c|}{\textbf{TUSZ}} &
  \multicolumn{2}{c}{\multirow{2}{*}{\textbf{Average}}} \\ \cmidrule{2-9}
\multicolumn{1}{c|}{} &
  \multicolumn{2}{c}{\textbf{4 Classes}} &
  \multicolumn{2}{c|}{\textbf{5 Classes}} &
  \multicolumn{2}{c}{\textbf{2 Classes}} &
  \multicolumn{2}{c|}{\textbf{4 Classes}} &
  \multicolumn{2}{c}{} \\
\multicolumn{1}{c|}{} &
  \textbf{Acc.} &
  \textbf{F1} &
  \textbf{Acc.} &
  \multicolumn{1}{c|}{\textbf{F1}} &
  \textbf{Acc.} &
  \textbf{F1} &
  \textbf{Acc.} &
  \multicolumn{1}{c|}{\textbf{F1}} &
  \textbf{Acc.} &
  \textbf{F1} \\ \midrule
\textbf{MedualTime (BERT)}         & 0.83 & 0.76 & 0.81 & 0.74 & 0.84 & 0.69 & 0.79 & 0.78 & 0.82          & \textbf{0.74} \\
\textbf{MedualTime (RoBERTa)}      & 0.83 & 0.76 & 0.80 & 0.73 & 0.87 & 0.61 & 0.79 & 0.74 & 0.82          & 0.71          \\
\textbf{MedualTime (GPT-2)}        & 0.82 & 0.75 & 0.80 & 0.73 & 0.86 & 0.52 & 0.72 & 0.60 & 0.80          & 0.65          \\
\textbf{MedualTime (ClinicalBERT)} & 0.83 & 0.76 & 0.81 & 0.75 & 0.87 & 0.68 & 0.79 & 0.75 & \textbf{0.83} & 0.73

            \\ 
            \bottomrule 
            \end{tabular} 
            
        
        \end{threeparttable} 
        
        } 

\caption{Textual Encoders Analysis under supervised learning.} 
\label{tab:enc} 
\end{table} 

\textbf{(b) Textual Encoder Analysis}
We investigate the impact of various textual encoders by examining the following options: BERT, RoBERTa \cite{liu2019roberta}, ClinicalBERT \cite{wang2023optimized}, and GPT-2. The results of supervised learning are in Table \ref{tab:enc}. More results about unsupervised learning are in \textbf{Appendix 1.5}. As shown in Table \ref{tab:enc}, BERT-based textual encoders consistently outperform GPT-2. This is likely due to GPT-2's primary focus on text generation, while BERT and its variants excel in comprehending the entire textual input. ClinicalBERT, pre-trained on medical texts, achieves the highest accuracy on average, perhaps attributed to its domain-specific pre-training knowledge. 

\begin{figure}[htb]
  \centering
  \includegraphics[width=0.48\textwidth]{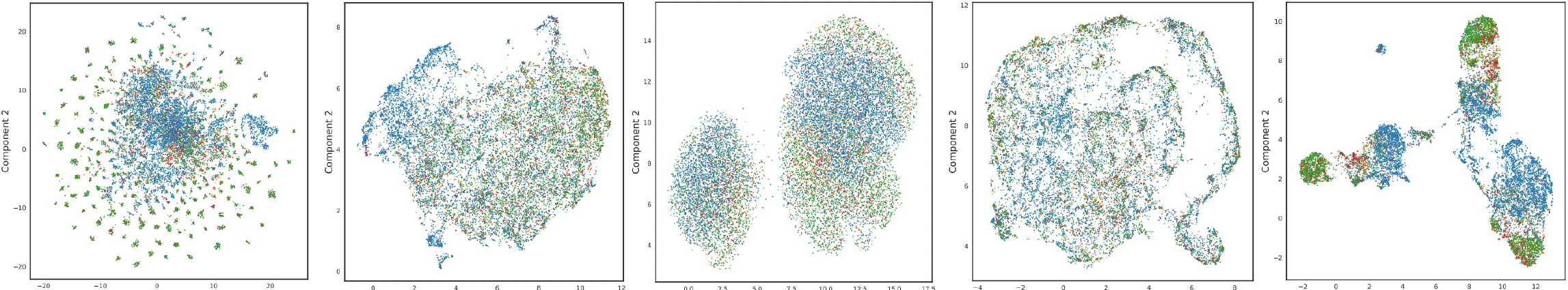}
    \caption{Embedding visualizations of different encoders on PTB-XL (4 Classes), i.e. ClinicalBERT, TS2Vec, MedualTime (Text), MedualTime (Time), MedualTime, from left to right. Different colors represent different labels.}
  \label{fig:ucm}
  \vspace{-3mm}
\end{figure}

    

\textbf{\textbf{(c) Visualization}} To better visualize the learned representations of unsupervised learning, we use UMAP \cite{mcinnes2018umap} to project the unsupervised representation learning results into 2D plots. \textbf{(1)} Figure \ref{fig:ucm} displays the embeddings of various encoders on PTB-XL, with labels assigned to different categories. Both TS2Vec (time-only) and ClinicalBERT (text-only) cannot successfully identify different categories of ECGs.
\textbf{(2)} Compared with ClinicalBERT, MedualTime (Text) can better distinguish abnormal ECG and normal ECG, indicating the effectiveness of two modalities over one modality. \textbf{(3)} Compared with MedualTime (Time), MedualTime (Text) has obviously better discriminative capacity, supporting the advantage of textual-primary modeling over temporal-primary modeling. \textbf{(4)} Overall, MedualTime provides the most distinct representations, attributed to the benefits of complementary multimodal modeling.

\begin{figure}[htb]
  \vspace{-3mm}
  \begin{center}
    \includegraphics[width=0.45\textwidth]{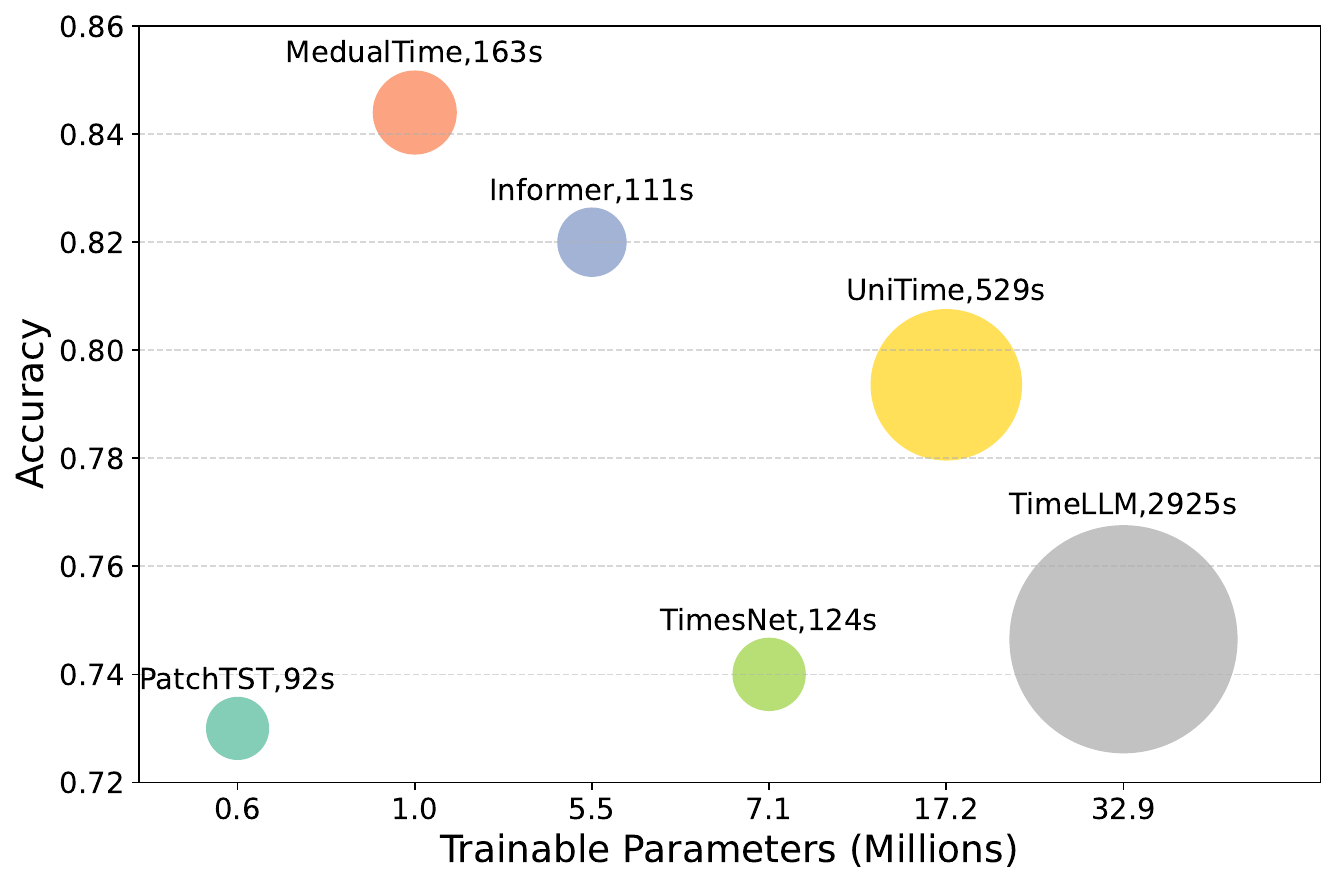} 
  \end{center}
  \vspace{-4mm}
  \caption{Efficiency comparison on the TUSZ dataset (2 Classes). The bubble size represents the relative training time per epoch.}
    \label{fig:efficiency}
    \vspace{-4mm}
\end{figure}

\textbf{(d) Efficiency Evaluation}
Due to limited space, we compare the computational costs in seconds on representative baselines and our model under supervised setting in Figure \ref{fig:efficiency}. MedualTime employs a frozen LLM backbone and introduces learnable adaptation tokens, enabling more efficient fine-tuning. It has approximately 1.0 million trainable parameters, even smaller than TimesNet with 7.1 M. Its training efficiency (163s per epoch) is close to TimesNet (124s per epoch) even though it has far larger total parameters. Compared with UniTime and TimeLLM,  MedualTime has the shortest training time and highest accuracy, highlighting its efficiency and superior performance. 

\section{Conclusion}
\label{sec:con}
\label{sec:Conclusions}
We propose a new textual-temporal paradigm for medical time series-text multimodal learning to explore the complementary modeling of different modalities. Our model leverages a dual-adapter design for temporal-primary and textual-primary modeling, with high-level multimodal fusion achieved through learnable token injection in the top layers of a shared frozen pre-trained LM pipeline, ensuring lightweight fine-tuning. Extensive experiments demonstrate that our model significantly enhances representation learning of medical data in both supervised and unsupervised settings. Few-shot label transfer experiments highlight its transferability. Our model leverages the capabilities of language models to equip clinicians with multimodal tools that enhance medical diagnosis.


\section*{Acknowledgments}
This work is funded by National Natural Science Foundation of China Grant No. 72371217, the Guangzhou Industrial Informatic and Intelligence Key Laboratory No. 2024A03J0628, the Nansha Key Area Science and Technology Project No. 2023ZD003, and Project No. 2021JC02X191.
\section*{Contribution Statement}
$^\dagger$ Co-first Authors, $^*$ Corresponding Authors.



\bibliographystyle{named}
\bibliography{ref}

\begin{thebibliography}{}

\bibitem[\protect\citeauthoryear{Chan \bgroup \em et al.\egroup }{2024}]{chan2024medtsllm}
Nimeesha Chan, Felix Parker, William Bennett, Tianyi Wu, Mung~Yao Jia, James Fackler, and Kimia Ghobadi.
\newblock Medtsllm: Leveraging llms for multimodal medical time series analysis.
\newblock {\em arXiv preprint arXiv:2408.07773}, 2024.

\bibitem[\protect\citeauthoryear{Chang \bgroup \em et al.\egroup }{2023}]{LLM4TS}
Ching Chang, Wen-Chih Peng, and Tien-Fu Chen.
\newblock {{LLM4TS}}: {{Two-Stage Fine-Tuning}} for {{Time-Series Forecasting}} with {{Pre-Trained LLMs}}.
\newblock In {\em arXiv:2308.08469}, 2023.

\bibitem[\protect\citeauthoryear{Cheng \bgroup \em et al.\egroup }{2024}]{cheng2024advancing}
Mingyue Cheng, Yiheng Chen, Qi~Liu, Zhiding Liu, and Yucong Luo.
\newblock Advancing time series classification with multimodal language modeling.
\newblock {\em arXiv preprint arXiv:2403.12371}, 2024.

\bibitem[\protect\citeauthoryear{Davuluri}{2024}]{davuluri2024overview}
Manaswini Davuluri.
\newblock An overview of natural language processing in analyzing clinical text data for patient health insights.
\newblock {\em Research-gate journal}, 10(10), 2024.

\bibitem[\protect\citeauthoryear{Devlin \bgroup \em et al.\egroup }{2018}]{devlin2018bert}
Jacob Devlin, Ming-Wei Chang, Kenton Lee, and Kristina Toutanova.
\newblock Bert: Pre-training of deep bidirectional transformers for language understanding.
\newblock {\em arXiv preprint arXiv:1810.04805}, 2018.

\bibitem[\protect\citeauthoryear{Eldele \bgroup \em et al.\egroup }{2021}]{tstcc}
Emadeldeen Eldele, Mohamed Ragab, Zhenghua Chen, Min Wu, Chee~Keong Kwoh, Xiaoli Li, and Cuntai Guan.
\newblock Time-series representation learning via temporal and contextual contrasting.
\newblock In {\em Proceedings of the Thirtieth International Joint Conference on Artificial Intelligence, {IJCAI-21}}, pages 2352--2359, 2021.

\bibitem[\protect\citeauthoryear{Fries \bgroup \em et al.\egroup }{2021}]{fries2021ontology}
Jason~A Fries, Ethan Steinberg, Saelig Khattar, Scott~L Fleming, Jose Posada, Alison Callahan, and Nigam~H Shah.
\newblock Ontology-driven weak supervision for clinical entity classification in electronic health records.
\newblock {\em Nature communications}, 12(1):2017, 2021.

\bibitem[\protect\citeauthoryear{Gruver \bgroup \em et al.\egroup }{2024}]{LLMTIME}
Nate Gruver, Marc Finzi, Shikai Qiu, and Andrew~G Wilson.
\newblock Large language models are zero-shot time series forecasters.
\newblock {\em Advances in Neural Information Processing Systems}, 36, 2024.

\bibitem[\protect\citeauthoryear{Guo \bgroup \em et al.\egroup }{2019}]{guo2019deep}
Wenzhong Guo, Jianwen Wang, and Shiping Wang.
\newblock Deep multimodal representation learning: A survey.
\newblock {\em Ieee Access}, 7:63373--63394, 2019.

\bibitem[\protect\citeauthoryear{Hochreiter and Schmidhuber}{1997}]{hochreiter1997lstm}
Sepp Hochreiter and J{\"u}rgen Schmidhuber.
\newblock Long short-term memory.
\newblock {\em Neural computation}, 9(8):1735--1780, 1997.

\bibitem[\protect\citeauthoryear{Jia \bgroup \em et al.\egroup }{2024}]{jia2024gpt4mts}
Furong Jia, Kevin Wang, Yixiang Zheng, Defu Cao, and Yan Liu.
\newblock Gpt4mts: Prompt-based large language model for multimodal time-series forecasting.
\newblock In {\em Proceedings of the AAAI Conference on Artificial Intelligence}, volume~38, pages 23343--23351, 2024.

\bibitem[\protect\citeauthoryear{Jin \bgroup \em et al.\egroup }{2023}]{jin2023timellm}
Ming Jin, Shiyu Wang, Lintao Ma, Zhixuan Chu, James~Y Zhang, Xiaoming Shi, Pin-Yu Chen, Yuxuan Liang, Yuan-Fang Li, Shirui Pan, et~al.
\newblock Time-llm: Time series forecasting by reprogramming large language models.
\newblock In {\em The Twelfth International Conference on Learning Representations}, 2023.

\bibitem[\protect\citeauthoryear{King \bgroup \em et al.\egroup }{2023}]{king2023multimodal}
Ryan King, Tianbao Yang, and Bobak~J Mortazavi.
\newblock Multimodal pretraining of medical time series and notes.
\newblock In {\em Machine Learning for Health (ML4H)}, pages 244--255. PMLR, 2023.

\bibitem[\protect\citeauthoryear{Kingma}{2014}]{kingma2014adam}
Diederik~P Kingma.
\newblock Adam: A method for stochastic optimization.
\newblock {\em arXiv preprint arXiv:1412.6980}, 2014.

\bibitem[\protect\citeauthoryear{Kitaev \bgroup \em et al.\egroup }{2020}]{kitaev2020reformer}
Nikita Kitaev, {\L}ukasz Kaiser, and Anselm Levskaya.
\newblock Reformer: The efficient transformer.
\newblock {\em arXiv preprint arXiv:2001.04451}, 2020.

\bibitem[\protect\citeauthoryear{Li \bgroup \em et al.\egroup }{2024}]{li2024frozen}
Jun Li, Che Liu, Sibo Cheng, Rossella Arcucci, and Shenda Hong.
\newblock Frozen language model helps ecg zero-shot learning.
\newblock In {\em Medical Imaging with Deep Learning}, pages 402--415. PMLR, 2024.

\bibitem[\protect\citeauthoryear{Liang \bgroup \em et al.\egroup }{2024}]{liang2024foundations}
Paul~Pu Liang, Amir Zadeh, and Louis-Philippe Morency.
\newblock Foundations \& trends in multimodal machine learning: Principles, challenges, and open questions.
\newblock {\em ACM Computing Surveys}, 56(10):1--42, 2024.

\bibitem[\protect\citeauthoryear{Liu \bgroup \em et al.\egroup }{}]{liuzero}
Che Liu, Zhongwei Wan, Cheng Ouyang, Anand Shah, Wenjia Bai, and Rossella Arcucci.
\newblock Zero-shot ecg classification with multimodal learning and test-time clinical knowledge enhancement.
\newblock In {\em Forty-first International Conference on Machine Learning}.

\bibitem[\protect\citeauthoryear{Liu \bgroup \em et al.\egroup }{2021}]{liu2021pyraformer}
Shizhan Liu, Hang Yu, Cong Liao, Jianguo Li, Weiyao Lin, Alex~X Liu, and Schahram Dustdar.
\newblock Pyraformer: Low-complexity pyramidal attention for long-range time series modeling and forecasting.
\newblock In {\em International conference on learning representations}, 2021.

\bibitem[\protect\citeauthoryear{Liu \bgroup \em et al.\egroup }{2023}]{liu2023itransformer}
Yong Liu, Tengge Hu, Haoran Zhang, Haixu Wu, Shiyu Wang, Lintao Ma, and Mingsheng Long.
\newblock itransformer: Inverted transformers are effective for time series forecasting.
\newblock {\em arXiv preprint arXiv:2310.06625}, 2023.

\bibitem[\protect\citeauthoryear{Liu \bgroup \em et al.\egroup }{2024}]{liu2023unitime}
Xu~Liu, Junfeng Hu, Yuan Li, Shizhe Diao, Yuxuan Liang, Bryan Hooi, and Roger Zimmermann.
\newblock Unitime: A language-empowered unified model for cross-domain time series forecasting.
\newblock In {\em Proceedings of the ACM Web Conference 2024}, 2024.

\bibitem[\protect\citeauthoryear{Liu}{2019}]{liu2019roberta}
Yinhan Liu.
\newblock Roberta: A robustly optimized bert pretraining approach.
\newblock {\em arXiv preprint arXiv:1907.11692}, 2019.

\bibitem[\protect\citeauthoryear{McInnes \bgroup \em et al.\egroup }{2018}]{mcinnes2018umap}
Leland McInnes, John Healy, and James Melville.
\newblock Umap: Uniform manifold approximation and projection for dimension reduction.
\newblock {\em arXiv preprint arXiv:1802.03426}, 2018.

\bibitem[\protect\citeauthoryear{Nie \bgroup \em et al.\egroup }{2022}]{nie2022patchtst}
Yuqi Nie, Nam~H Nguyen, Phanwadee Sinthong, and Jayant Kalagnanam.
\newblock A time series is worth 64 words: Long-term forecasting with transformers.
\newblock {\em arXiv preprint arXiv:2211.14730}, 2022.

\bibitem[\protect\citeauthoryear{Radford \bgroup \em et al.\egroup }{2019}]{radford2019gpt2}
Alec Radford, Jeffrey Wu, Rewon Child, David Luan, Dario Amodei, Ilya Sutskever, et~al.
\newblock Language models are unsupervised multitask learners.
\newblock {\em OpenAI blog}, 1(8):9, 2019.

\bibitem[\protect\citeauthoryear{Shah \bgroup \em et al.\egroup }{2018}]{shah2018temple}
Vinit Shah, Eva Von~Weltin, Silvia Lopez, James~Riley McHugh, Lillian Veloso, Meysam Golmohammadi, Iyad Obeid, and Joseph Picone.
\newblock The temple university hospital seizure detection corpus.
\newblock {\em Frontiers in neuroinformatics}, 12:83, 2018.

\bibitem[\protect\citeauthoryear{Wagner \bgroup \em et al.\egroup }{2020}]{wagner2020ptb}
Patrick Wagner, Nils Strodthoff, Ralf-Dieter Bousseljot, Dieter Kreiseler, Fatima~I Lunze, Wojciech Samek, and Tobias Schaeffter.
\newblock Ptb-xl, a large publicly available electrocardiography dataset.
\newblock {\em Scientific data}, 7(1):1--15, 2020.

\bibitem[\protect\citeauthoryear{Wang \bgroup \em et al.\egroup }{2023}]{wang2023optimized}
Guangyu Wang, Xiaohong Liu, Zhen Ying, Guoxing Yang, Zhiwei Chen, Zhiwen Liu, Min Zhang, Hongmei Yan, Yuxing Lu, Yuanxu Gao, et~al.
\newblock Optimized glycemic control of type 2 diabetes with reinforcement learning: a proof-of-concept trial.
\newblock {\em Nature Medicine}, 29(10):2633--2642, 2023.

\bibitem[\protect\citeauthoryear{Wang \bgroup \em et al.\egroup }{2024}]{wang2024contrast}
Yihe Wang, Yu~Han, Haishuai Wang, and Xiang Zhang.
\newblock Contrast everything: A hierarchical contrastive framework for medical time-series.
\newblock {\em Advances in Neural Information Processing Systems}, 36, 2024.

\bibitem[\protect\citeauthoryear{Weimann and Conrad}{2021}]{weimann2021transfer}
Kuba Weimann and Tim~OF Conrad.
\newblock Transfer learning for ecg classification.
\newblock {\em Scientific reports}, 11(1):5251, 2021.

\bibitem[\protect\citeauthoryear{Woo \bgroup \em et al.\egroup }{2022}]{woo2022etsformer}
Gerald Woo, Chenghao Liu, Doyen Sahoo, Akshat Kumar, and Steven Hoi.
\newblock Etsformer: Exponential smoothing transformers for time-series forecasting.
\newblock {\em arXiv preprint arXiv:2202.01381}, 2022.

\bibitem[\protect\citeauthoryear{Wu \bgroup \em et al.\egroup }{2021}]{wu2021autoformer}
Haixu Wu, Jiehui Xu, Jianmin Wang, and Mingsheng Long.
\newblock Autoformer: Decomposition transformers with auto-correlation for long-term series forecasting.
\newblock {\em Advances in neural information processing systems}, 34:22419--22430, 2021.

\bibitem[\protect\citeauthoryear{Wu \bgroup \em et al.\egroup }{2022}]{wu2022timesnet}
Haixu Wu, Tengge Hu, Yong Liu, Hang Zhou, Jianmin Wang, and Mingsheng Long.
\newblock Timesnet: Temporal 2d-variation modeling for general time series analysis.
\newblock In {\em The eleventh international conference on learning representations}, 2022.

\bibitem[\protect\citeauthoryear{Xie \bgroup \em et al.\egroup }{2022}]{9845479}
Jin Xie, Jie Zhang, Jiayao Sun, Zheng Ma, Liuni Qin, Guanglin Li, Huihui Zhou, and Yang Zhan.
\newblock A transformer-based approach combining deep learning network and spatial-temporal information for raw eeg classification.
\newblock {\em IEEE Transactions on Neural Systems and Rehabilitation Engineering}, 30:2126--2136, 2022.

\bibitem[\protect\citeauthoryear{Yu \bgroup \em et al.\egroup }{2024}]{yu2024ecg}
Han Yu, Peikun Guo, and Akane Sano.
\newblock Ecg semantic integrator (esi): A foundation ecg model pretrained with llm-enhanced cardiological text.
\newblock {\em arXiv preprint arXiv:2405.19366}, 2024.

\bibitem[\protect\citeauthoryear{Yue \bgroup \em et al.\egroup }{2022}]{yue2022ts2vec}
Zhihan Yue, Yujing Wang, Juanyong Duan, Tianmeng Yang, Congrui Huang, Yunhai Tong, and Bixiong Xu.
\newblock Ts2vec: Towards universal representation of time series.
\newblock In {\em Proceedings of the AAAI Conference on Artificial Intelligence}, volume~36, pages 8980--8987, 2022.

\bibitem[\protect\citeauthoryear{Zeng \bgroup \em et al.\egroup }{2023}]{zeng2023dlinear}
Ailing Zeng, Muxi Chen, Lei Zhang, and Qiang Xu.
\newblock Are transformers effective for time series forecasting?
\newblock In {\em Proceedings of the AAAI conference on artificial intelligence}, volume~37, pages 11121--11128, 2023.

\bibitem[\protect\citeauthoryear{Zhang and Yan}{2022}]{zhang2022crossformer}
Yunhao Zhang and Junchi Yan.
\newblock Crossformer: Transformer utilizing cross-dimension dependency for multivariate time series forecasting.
\newblock In {\em The eleventh international conference on learning representations}, 2022.

\bibitem[\protect\citeauthoryear{Zhang \bgroup \em et al.\egroup }{2022}]{zhang2022lightts}
Tianping Zhang, Yizhuo Zhang, Wei Cao, Jiang Bian, Xiaohan Yi, Shun Zheng, and Jian Li.
\newblock Less is more: Fast multivariate time series forecasting with light sampling-oriented mlp structures.
\newblock {\em arXiv preprint arXiv:2207.01186}, 2022.

\bibitem[\protect\citeauthoryear{Zhang \bgroup \em et al.\egroup }{2023a}]{zhang2023llama}
Renrui Zhang, Jiaming Han, Chris Liu, Peng Gao, Aojun Zhou, Xiangfei Hu, Shilin Yan, Pan Lu, Hongsheng Li, and Yu~Qiao.
\newblock Llama-adapter: Efficient fine-tuning of language models with zero-init attention.
\newblock {\em arXiv preprint arXiv:2303.16199}, 2023.

\bibitem[\protect\citeauthoryear{Zhang \bgroup \em et al.\egroup }{2023b}]{zhang2023co}
Weiqi Zhang, Jianfeng Zhang, Jia Li, and Fugee Tsung.
\newblock A co-training approach for noisy time series learning.
\newblock In {\em Proceedings of the 32nd ACM International Conference on Information and Knowledge Management}, pages 3308--3318, 2023.

\bibitem[\protect\citeauthoryear{Zhou \bgroup \em et al.\egroup }{2021}]{zhou2021informer}
Haoyi Zhou, Shanghang Zhang, Jieqi Peng, Shuai Zhang, Jianxin Li, Hui Xiong, and Wancai Zhang.
\newblock Informer: Beyond efficient transformer for long sequence time-series forecasting.
\newblock In {\em Proceedings of the AAAI conference on artificial intelligence}, volume~35, pages 11106--11115, 2021.

\bibitem[\protect\citeauthoryear{Zhou \bgroup \em et al.\egroup }{2022}]{zhou2022fedformer}
Tian Zhou, Ziqing Ma, Qingsong Wen, Xue Wang, Liang Sun, and Rong Jin.
\newblock Fedformer: Frequency enhanced decomposed transformer for long-term series forecasting.
\newblock In {\em International conference on machine learning}, pages 27268--27286. PMLR, 2022.

\bibitem[\protect\citeauthoryear{Zhou \bgroup \em et al.\egroup }{2024}]{zhou2024one}
Tian Zhou, Peisong Niu, Liang Sun, Rong Jin, et~al.
\newblock One fits all: Power general time series analysis by pretrained lm.
\newblock {\em Advances in neural information processing systems}, 36, 2024.

\end{thebibliography}


\end{document}